\documentclass{article} %
\PassOptionsToPackage{table}{xcolor}
\usepackage{iclr2027_conference,times}

\usepackage{amsmath,amsfonts,bm}

\def\eqref#1{equation~\ref{#1}}

\def\1{\bm{1}}

\def\va{{\bm{a}}}

\def\vj{{\bm{j}}}

\def\vm{{\bm{m}}}

\def\vp{{\bm{p}}}

\def\mA{{\bm{A}}}

\DeclareMathAlphabet{\mathsfit}{\encodingdefault}{\sfdefault}{m}{sl}
\SetMathAlphabet{\mathsfit}{bold}{\encodingdefault}{\sfdefault}{bx}{n}

\usepackage{hyperref}
\usepackage[capitalise,noabbrev]{cleveref}
\usepackage{url}
\usepackage{graphicx}
\usepackage{booktabs}
\usepackage{multirow}
\usepackage{placeins}
\usepackage{wrapfig}
\usepackage{amsthm}

\usepackage{xcolor}
\usepackage{tcolorbox}
\usepackage{fontawesome5}
\tcbuselibrary{skins,breakable}

\definecolor{tokenwise}{HTML}{238B45}   \definecolor{tokenwiselight}{HTML}{74C476}
\definecolor{headwise}{HTML}{6A51A3}    \definecolor{headwiselight}{HTML}{9E9AC8}
\definecolor{attention}{HTML}{6BAED6}
\definecolor{influence}{HTML}{08519C}   %
\definecolor{agreementink}{HTML}{2A78D6}   %
\definecolor{lossink}{HTML}{EB6834}        %

\newif\ifdefcolors \defcolorstrue     %
\newcommand{\tw}[1]{\ifdefcolors{\color{tokenwise}#1}\else{#1}\fi}   %
\newcommand{\hw}[1]{\ifdefcolors{\color{headwise}#1}\else{#1}\fi}    %
\newcommand{\iw}[1]{\ifdefcolors{\color{influence}#1}\else{#1}\fi}   %
\newcommand{\headmass}{Head Mass}  %
\newcommand{\agreementink}[1]{\ifdefcolors{\color{agreementink}#1}\else{#1}\fi}
\newcommand{\lossink}[1]{\ifdefcolors{\color{lossink}#1}\else{#1}\fi}

\newcommand{\marg}[1]{\vm^{(#1)}}              %
\newcommand{\margel}[2]{m^{(#1)}_{#2}}         %
\newcommand{\wt}[3]{\omega^{(#1)}_{#2,#3}}     %
\newcommand{\attn}[1]{\mA^{(#1)}}              %
\newcommand{\attnel}[3]{A^{(#1)}_{#2,#3}}      %
\newif\ifcolargs \colargstrue         %
\makeatletter
\newcommand{\colterm}[2]{\@ifnextchar'{\colterm@prime{#1}{#2}}{\colterm@arg{#1}{#2}}}
\def\colterm@prime#1#2'{\colterm@arg{#1}{#2'}}
\def\colterm@arg#1#2{\@ifnextchar({\colterm@paren{#1}{#2}}{#1{#2}}}
\def\colterm@paren#1#2(#3){\ifcolargs#1{#2(#3)}\else#1{#2}(#3)\fi}
\makeatother
\newcommand{\tokw}{\colterm\tw{\va}}           %
\newcommand{\tokwel}[1]{\colterm\tw{a_{#1}}}   %
\newcommand{\headw}{\colterm\hw{\vp}}          %
\newcommand{\headwel}[1]{\colterm\hw{p_{#1}}}  %
\newcommand{\infl}{\colterm\iw{\vj}}           %
\newcommand{\inflel}[1]{\colterm\iw{j_{#1}}}   %

\newtheoremstyle{boxedthm}{0pt}{0pt}{\itshape}{}{\bfseries}{.}{.5em}{}
\theoremstyle{boxedthm}
\newtheorem{definition}{Definition}
\newtheorem{lemma}{Lemma}
\newtheorem{assumption}{Assumption}
\newtheorem{theorem}{Theorem}
\newcommand{\restatedthmnum}{}
\newtheorem*{restatedtheoreminner}{Theorem~\restatedthmnum}
\newenvironment{restatedtheorem}[1]
  {\renewcommand{\restatedthmnum}{\ref{#1}}\restatedtheoreminner}
  {\endrestatedtheoreminner}
\theoremstyle{plain}

\definecolor{thmrule}{RGB}{128, 128, 128}  %
\definecolor{thmback}{RGB}{245, 245, 245}  %
\definecolor{thmtext}{RGB}{ 40,  40,  40}  %
\tcbset{
  theoremstyle/.style={
    enhanced,
    breakable,
    sharp corners,
    colback=thmback,
    colframe=thmrule,
    coltext=thmtext,
    boxrule=0pt,
    leftrule=1.5pt,
    boxsep=0pt,
    left=6pt,
    right=6pt,
    top=3pt,
    bottom=3pt,
    before skip=6pt plus 2pt,
    after skip=6pt plus 2pt,
    before upper={%
      \setlength{\abovedisplayskip}{4pt plus 1pt}%
      \setlength{\belowdisplayskip}{4pt plus 1pt}%
      \setlength{\abovedisplayshortskip}{2pt plus 1pt}%
      \setlength{\belowdisplayshortskip}{2pt plus 1pt}%
    },
  }
}

\tcolorboxenvironment{definition}{theoremstyle}
\tcolorboxenvironment{lemma}{theoremstyle}
\tcolorboxenvironment{assumption}{theoremstyle}
\tcolorboxenvironment{theorem}{theoremstyle}
\tcolorboxenvironment{restatedtheorem}{theoremstyle}

\newcommand{\takeaway}[1]{%
  \begin{tcolorbox}[theoremstyle]%
    {\color{thmrule}\faLightbulb[regular]}\enspace\textbf{Takeaway.}\enspace #1%
  \end{tcolorbox}%
}

\usepackage{enumitem}
\setlist[itemize]{topsep=2pt,itemsep=1pt,parsep=0pt,leftmargin=*}
\makeatletter
\g@addto@macro\normalsize{%
  \setlength{\abovedisplayskip}{5pt plus 2pt minus 1pt}%
  \setlength{\belowdisplayskip}{5pt plus 2pt minus 1pt}%
  \setlength{\abovedisplayshortskip}{2pt plus 1pt}%
  \setlength{\belowdisplayshortskip}{3pt plus 1pt}}
\makeatother

\crefname{assumption}{Assumption}{Assumptions}
\Crefname{assumption}{Assumption}{Assumptions}

\usepackage{array}
\definecolor{kvshade}{HTML}{80CBC4}
\definecolor{kvoffline}{HTML}{B07326}
\definecolor{kvonline}{HTML}{4C6B8A}

\title{Emergent Structure in the Marginal \\Attention Space of Language Models}

\author{\parbox[t]{\dimexpr\textwidth-2\tabcolsep\relax}{\centering
Valentino Maiorca\thanks{Equal contribution.} \qquad Walter Nelson\footnotemark[1] \qquad Francesco Locatello \\[0.5em]
\mdseries Institute of Science and Technology Austria (ISTA) \\
\texttt{<firstname>.<lastname>@ista.ac.at}
}}

\iclrpreprint %
\begin{document}

\maketitle

\begin{abstract}
While representation similarity across independently trained language models is well-documented, how internal mechanics such as attention behave across models remains far less characterized. 
Inspired by this gap, we examine the structure of post-softmax attention weights by marginalizing over query positions, mapping them into a joint token--head ``marginal attention space''.
Evaluating across 60+ diverse LLMs, we find that different properties emerge when reducing this space along its token and head axes.
When reduced \textit{token-wise}, marginal attention yields a text-intrinsic signal robustly conserved \textit{across models}.
To explain this property, we empirically connect marginal attention to the input--output Jacobian of the network, and prove theoretically that under a smoothness assumption, models with similar next-token distributions are guaranteed to have similar input--output Jacobian statistics. 
When reduced \textit{head-wise}, it forms a model-private signature conserved \textit{across documents}. Practically, this provides a natural way to estimate a per-head budget for key-value (KV) cache eviction, effectively decoupling model-specific budget allocation from text-intrinsic token scoring. 
On standard eviction benchmarks, a per-head budget precomputed offline on pretraining text, combined with a training-free token score, shows competitive performance with methods that recompute the budget on every document or train it per target.
Code available at \url{https://github.com/Flegyas/marginal-attention}.

\end{abstract}

\section{Introduction}
\label{sec:intro}
Despite fundamental differences in architecture, initialization and training objective, neural networks have repeatedly been shown to rediscover a \textit{similar representational geometry}: the embeddings two models assign to semantically related inputs agree up to a simple transformation \citep{pmlr-v97-kornblith19a,svcca2017,morcos2018insights}. This phenomenon is strong enough that unrelated models can be composed \citep{Lenc_2015_CVPR,ranzato2021stitching}, even across different architectures, data or modalities \citep{moschella2023relative,norelli2022,maiorca2023}. Building on this empirical evidence, recent work has turned from \emph{whether} representations agree to \emph{why} they do, with competing hypotheses \citep{pmlr-v235-huh24a,groger2026revisiting}.

\begin{figure}[t]
    \centering
    \includegraphics[width=\linewidth]{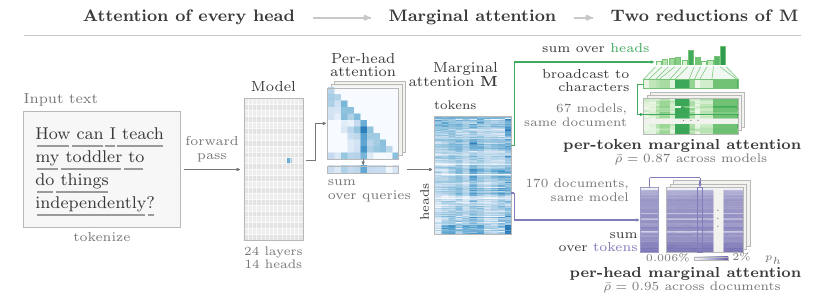}
    \input{figures/teaser_caption.tex}
    \label{fig:teaser}
\end{figure}

How the \textit{mechanics} that produce these representations behave across models, attention in transformers above all, is far less characterized, and is the focus of this paper.
We study attention through its \textit{marginals}: the attention each head pays to an input token, weighted by what the head writes from it and averaged over output positions.
On one document, this gives a matrix with one row per head and one column per input token, which we reduce along each axis (\cref{fig:teaser}).
Reducing over heads gives the \textit{\tw{per-token marginal attention}}, a property of the text, shared across models that differ in lab, data, size, tokenizer and architecture.
Reducing over tokens gives the \textit{\hw{per-head marginal attention}}, a property of the model, stable across documents and not shared across unrelated models.

To explain the \tw{per-token} observation, we connect marginal attention to the input-output Jacobian of the language model. In particular, we show that marginal attention, although a simple statistic of the attention matrices alone, approximates the blockwise norms of the input-output Jacobian well. The attention paid to input position $s$ by an output position $t$, averaged across all heads in the model, closely approximates the dependence of the output logits at $t$ on the input token $s$ given by the Jacobian. This matters because the Jacobian is a local description of the language model's entire computation, which includes the value stream and higher-order attention dependencies as well as first-order attention effects.
Our main theoretical result shows that two smooth models with nearly the same next-token distributions must have similar input-output Jacobians. For models that share a tokenizer the theory is predictive: pairs with closer next-token predictions agree more, in both the marginal influence and the \tw{per-token marginal attention}.

The \hw{per-head} observation shows that our quantity is a reliable model-specific signal that is largely independent of the input document, and we leverage this for key-value (KV) cache eviction. An eviction method must decide how much of the cache each head keeps, and it must keep what later queries will retrieve. We therefore build from our per-head marginal attention a static budget computed on the reconstruction queries of a second pass that asks the model to repeat the document \citep{NEURIPS2025_f4eaa4b8}. On those queries the profile is also stable across documents, so the budget is computed once per model, offline, on pretraining text. At inference only a per-token score remains, and since the per-token marginal attention is shared across models, that scoring can also be adjusted by a different, much smaller model.

Our main contributions can be summarized as follows:
\begin{itemize}
    \item We propose \textit{marginal attention} as a relevant quantity to study in language models. Aggregated over heads, it is shared across unrelated models. Per head, it is specific to each model.
    \item We connect attention statistics in language models to the input-output Jacobian, and show that model smoothness and next-token distribution similarity alone are theoretically sufficient for input-output Jacobian near-identifiability across models (\cref{thm:jacobianControl}).
    \item We show that \textit{\hw{per-head marginal attention}} is a stable quantity across documents, enabling efficient KV cache eviction via a one-time, document-agnostic head budget calculation.
\end{itemize}

\FloatBarrier
\section{Related work}
\label{sec:related}

\paragraph{Representation convergence.}
\label{sec:related:representation}
Recently, a considerable body of work has examined the similarity of representations across independently trained models, including large language models (LLMs). These works propose similarity metrics \citep{pmlr-v97-kornblith19a,svcca2017,morcos2018insights}, compose models \citep{Lenc_2015_CVPR,ranzato2021stitching,norelli2022,maiorca2023}, or study the geometry of the representation space \citep{moschella2023relative,pmlr-v235-huh24a}.
Here, we shift the focus from representation-level to the \textit{mechanics} that produce these representations, the interactions between data inside the model. In transformer-based language models this is largely determined by the attention mechanism and, as the next paragraph reviews, attention itself is not identifiable across models. In our work, we propose a marginal version of attention whose \tw{per-token} reduction is a property of the text, shared and conserved across models.

\paragraph{Attention compared across models.}
\label{sec:related:attention}
The attention mechanism is the core component of transformer-based models. 
On one hand, when treated as a free parameter, the attention matrix has been shown to be non-identifiable \citep{brunner2020id}, and specific heads are generally not empirically identifiable across models \citep{xu2026patternselectivitytaskcausalstructure,ICLR2026_3c1fe56b}, perhaps owing to their differing specializations \citep{voita2019analyzing,michel2019sixteenheads,textspan,residual,headpursuit}. 
On the other hand, summary statistics of attention maps have been observed to be well-conserved across models, with \citet{ola2025} first noting that they can be sufficient to differentiate documents visually. Other work exploits the emergent property of attention that it approximates the computational map for explanations \citep{abnar-zuidema-2020-quantifying} and model provenance assessment \citep{zhang2026attndiff}. In contrast, we show that marginal attention approximates the input--output Jacobian and prove that models with similar next-token predictions have similar Jacobians, which provides insight into why the \tw{per-token marginal attention} is near-identifiable across models.

\paragraph{KV cache eviction.}
\label{sec:related:eviction}
Attention has a well-known quadratic cost in the input length, and at inference this cost is paid in memory as well. Serving a long context requires storing the key and value vectors of every past token for every head. An active line of work, eviction, bounds that memory by keeping only a subset of positions inside a fixed budget, while alternatives compress every position instead \citep{monteiro2026kvkaizenlearningcontextadaptive} or learn a compact cache for a document by distillation \citep{laguna2026cartridgeskvcachecompression}. 
Most eviction methods rank positions by the attention they have received, keeping the sinks and a recent window \citep{ICLR2024_5e5fd18f}, the most attended positions \citep{NEURIPS2023_6ceefa7b,MLSYS2024_48fecef4,chari2025compactorcalibratedqueryagnostickv}, or those the current query attends to \citep{NEURIPS2024_28ab4182,zhang-etal-2025-clusterattn,298683}. A more recent line of works ranks a position by the attention it will receive, estimated from the queries seen so far \citep{devoto2025expected}, computed on a reconstruction pass \citep{NEURIPS2025_f4eaa4b8}, predicted by a trained surrogate \citep{jegou2026kvzapfastadaptivefaithful,kim2026fastkvzipefficientaccurate}, or learned directly as a policy over future utility \citep{moschella2026learningevictkeyvaluecache}. Head budgets follow from the observation that a few specialised heads perform long-context retrieval \citep{wu2024retrieval}, so heads are labelled offline as retrieval or streaming \citep{xiao2024duoattention,tang2024razorattention}, reallocated within a layer \citep{NEURIPS2025_a40ff56d} or over depth \citep{cai2025pyramidkvdynamickvcache}, or assigned a static table solved on a calibration context \citep{tang2026predictingfutureutilityglobal}. A smaller model of the same family has also been used in place of the target, by transplanting its attention \citep{zhao-etal-2025-iam} or lending its scores \citep{NEURIPS2025_45790132,li2026proxykvcrossmodelproxypruning}, only within a family; the cache itself has also been translated between models with trained maps \citep{dery2026latentspacecommunicationkv,castin2026kvlingolearningkvcachetranslators}. In this work, we build on the \hw{per-head marginal attention} to obtain a static head budget for eviction, and on the identifiability of marginal attention to adjust the token scorer with a smaller model.

\section{Marginal attention}
\label{sec:method}

\paragraph{Notation.}
Let $X = (x_1, \dots, x_n)$ be a sequence of $n$ tokens, and $f$ an autoregressive language model with $H$ attention heads indexed by $h \in [H] := \{1, \dots, H\}$. Throughout, $s$ indexes input positions and $t$ indexes output positions. Let $\attn{h}(X) \in [0,1]^{n \times n}$ denote the post-softmax attention matrix of head $h$, where $\attnel{h}{s}{t}(X)$ is the weight output position $t$ assigns to input position $s$. By causality, $\attnel{h}{s}{t}(X) = 0$ for $s > t$, and $\sum_{s=1}^{n} \attnel{h}{s}{t}(X) = 1$ for all $t$. We write $\Delta^{n-1}$ for the set of vectors with $n$ non-negative entries summing to one, i.e. the probability distributions over the $n$ positions.

The marginal attention of a head measures how much attention each input position receives from that head, averaged over the output positions.
\begin{definition}[Marginal attention]
\label{defn:marginalAttention}
Given weights $\wt{h}{s}{t}(X) \ge 0$, the \textbf{marginal attention} of head $h$ is the vector $\marg{h}(X) \in \mathbb{R}_{\ge 0}^{n}$ that averages weighted attention over output positions:
\begin{equation*}
    \margel{h}{s}(X) := \frac{1}{n} \sum_{t=1}^{n} \wt{h}{s}{t}(X)\, \attnel{h}{s}{t}(X).
\end{equation*}
\end{definition}

Throughout the paper we set $\omega$ to rescale head $h$'s attention to position $s$ by the norm of the value that position sends through the head, $\wt{h}{s}{t}(X) := \lVert W_O^{(h)} v_s(X) \rVert$, with $v_s(X)$ the value vector at position $s$ and $W_O^{(h)}$ head $h$'s slice of the output projection. The weight does not depend on the output position $t$, so $\margel{h}{s}(X)$ is head $h$'s average attention to position $s$ times the norm of what the head writes from that position. Both factors can be computed in a single forward pass. The unweighted variant, $\wt{h}{s}{t} \equiv 1$, gives similar agreement and is reported in \cref{app:agreement} as an ablation.

Stacked over the $H$ heads, the marginal attention of a document is a matrix with one row per head and one column per input position (\cref{fig:teaser}). We reduce it along each axis, with the same weights $\omega$ in both reductions. Averaging over heads gives the \textbf{\tw{per-token marginal attention}}, which measures how much attention each token receives from the model as a whole. 
\begin{definition}[Per-token marginal attention]
\label{defn:perToken}
The \textbf{\tw{per-token marginal attention}} is the sum of the heads' marginal attention, normalised to a distribution over positions:
\begin{equation*}
    \tokw(X) := \frac{1}{Z(X)} \sum_{h=1}^{H} \marg{h}(X) \;\in\; \Delta^{n-1},
\end{equation*}
where $Z(X) := \sum_{h=1}^{H} \sum_{s=1}^{n} \margel{h}{s}(X)$ is the total marginal attention of the document.
\end{definition}

\begin{figure}[ht]
\centering
\includegraphics[width=0.9\linewidth]{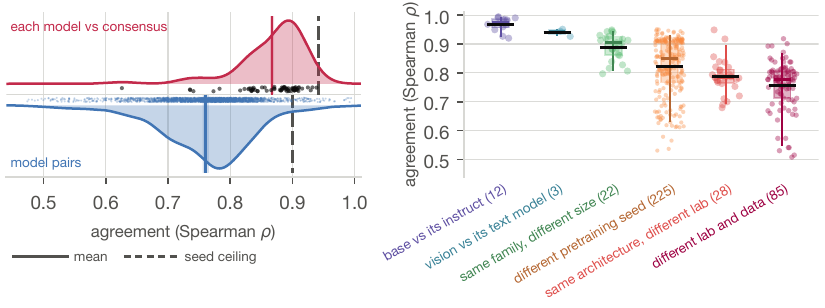}
\input{figures/main/census_distribution_caption.tex}
\label{fig:census_distribution}
\end{figure}

Summing over positions gives the \textbf{\hw{per-head marginal attention}}, which measures each head's share of the model's total marginal attention. 
\begin{definition}[Per-head marginal attention]
\label{defn:perHead}
The \textbf{\hw{per-head marginal attention}} over a set of documents $\mathcal{D}$ is each head's average share of the model's total marginal attention,
\begin{equation*}
    \headwel{h} := \mathbb{E}_{X\sim\mathcal{D}}\bigl[p_h(X)\bigr],
\end{equation*}
where $p_h(X) := W_h(X) / Z(X)$ and $W_h(X) := \sum_{s=1}^{n} \margel{h}{s}(X)$ is head $h$'s total marginal attention on document $X$. Thus $\headw \in \Delta^{H-1}$ is a single vector per model.
\end{definition}

\begin{figure}[ht]
\centering
\includegraphics[width=0.8\linewidth]{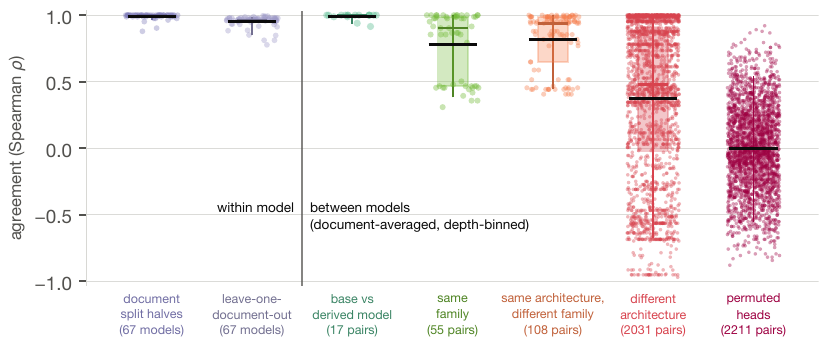}
\input{figures/main/head_write_distribution_caption.tex}
\label{fig:head_write_distribution}
\end{figure}

The per-token profile is indexed by the document's positions, so we compare it across models; the per-head profile is indexed by the model's heads, so we compare it across documents. We ask whether each reduction is a property of the text, shared across models, or a property of the model, stable across documents. In the following, we first define the two reductions (\cref{defn:perToken,defn:perHead}), show that the first is a property of the text and the second a property of the model (\cref{sec:census}), then explain the first theoretically through the Jacobian (\cref{sec:theory}), and finally use the second as a key-value cache budget (\cref{sec:application}).

\subsection{The two reductions across models and across documents}
\label{sec:census}

We now empirically test whether the \tw{per-token marginal attention} (\cref{defn:perToken}) is shared across models and the \hw{per-head marginal attention} (\cref{defn:perHead}) is stable across documents.

\paragraph{Experimental setting.}
\looseness-1
We compute both reductions on one set of models and documents, and compare each along two axes: across models on the same document, and across documents within the same model. We consider 67 open models from 38 families and 28 architectures, from 0.27B to 117B parameters, trained by different labs on different data with different tokenizers, including mixture-of-experts and multi-head-latent-attention models.
We use 850 natural documents, 50 from each of 17 domains of the Pile \citep{gao2020pile}; the \hw{per-head} comparison uses 170 of them, a seeded subsample of 10 per domain. Tokenizers differ, so for the \tw{per-token} profile we broadcast each token's value to the characters it covers, and compare two profiles per document by Spearman correlation after removing the average over 50 relative-position bins. Heads of different models have no correspondence, so we average each \hw{per-head} profile over documents and bin it into ten relative-depth bins before comparing two models; within one model we compare the profile averaged over one half of the documents with the profile averaged over the other half, on the same ten bins, and the profile of one document with the average over all the others, on the full vector over heads. As a reference for the \tw{per-token} profile, we compare each model with the consensus on a different document. We also show the agreement between ten pythia-410m reruns that differ only in the pretraining seed. As a reference for the \hw{per-head} profile, we permute each model's heads. Details are in \cref{app:census_details,app:agreement,app:head_details}.

\paragraph{Result analysis.}
\Cref{fig:census_distribution,fig:head_write_distribution} \textit{(left)} show the main comparison for each reduction. Across models, the agreement of each model's \tw{per-token} profile with the consensus of the other models is high (0.87), and its agreement with the consensus on a different document is near zero (0.02), so the agreement is specific to the document. Deleting punctuation and whitespace from the axis raises this agreement slightly, so it does not come from delimiters (\cref{tab:census_deletion}), and neither the number of models in the consensus nor the head weights change it much (\cref{app:agreement}).
Across documents, the \hw{per-head} profile one model computes on one half of the documents agrees closely with the profile it computes on the other half (0.99), and between models of different architectures agreement is much lower (0.37) on the same ten depth bins, so the profile is specific to the model. The right panels group model pairs by what they have in common; both profiles agree most between a model and a version trained from it and least between unrelated models (\cref{app:agreement:pairs,app:head_details}).

\takeaway{These results show that the \tw{per-token marginal attention} is a model-agnostic property of the text, and the \hw{per-head marginal attention} is a document-agnostic property of the model.}

\section{Identifiability in the marginal attention}
\label{sec:theory}

In this section, we explain the stability of \tw{per-token marginal attention} across models. We start with an empirical observation: across models, the \tw{per-token marginal attention} is reliably correlated with an analogous statistic of the \textit{input-output Jacobian}. In the second step, we show that this Jacobian statistic is a well-conserved quantity across models under mild assumptions (\cref{thm:jacobianControl}). We then validate this argument end-to-end by showing empirically that the agreement of two models' next-token logits predicts how well they agree on both the Jacobian statistic and the marginal attention. First, we define our Jacobian statistic and discuss a few of its basic properties. From here on, the elements of an input sequence $X$ are its token embeddings $\mathbf{x}_1, \dots, \mathbf{x}_n \in \mathbb{R}^d$, and $X$ is drawn from a distribution $\mathbf{P}(X)$ over sequences of any length $n$. A language model $f$ maps $X$ to one output logit vector per position, and its input-output Jacobian $J$ collects the derivatives of these logits with respect to the input embeddings; a second model $f'$ has Jacobian $J'$. We write $X^{(s) + r\mathbf{u}}$ for $X$ with the embedding at position $s$ moved by $r\mathbf{u}$, for a unit vector $\mathbf{u}$ and $0 \le r < R$, where $R > 0$ is the radius over which the assumptions of \cref{sec:theory:identifiability} hold.

\begin{definition}
    \label{defn:marginalInfluence}
    For an input sequence $X$ of length $n$ and a causal language model $f$ with input-output Jacobian $J$, we denote the \textbf{\iw{per-token marginal influence}} $\infl(X)$ at position $s$ as
    \begin{equation*}
        \inflel{s}(X) := \frac{\sum_{t \geq s} \left\lVert \left[ J^{(s)}(X) \right]_t \right\rVert_F}{\sum_{s',t} \left\lVert \left[ J^{(s')}(X) \right]_t \right\rVert_F}
    \end{equation*}
    where $s, s'$ index input positions and $t$ indexes output positions, so $[J^{(s)}]_t$ is the block of the Jacobian which locally linearizes the effect of input position $s$ on the logits at output position $t$.
\end{definition}
By causal masking, we have that $\infl(X) \in \Delta^{n-1}$, the $(n-1)$-simplex, so $\infl(X)$ can be seen as a distribution over input positions. Now that it is placed in the same space $\Delta^{n-1}$ as the \tw{per-token marginal attention} $\tokw(X)$ (\cref{defn:perToken}), we can establish the correlation between the two quantities.

\subsection{Marginal attention approximates the Jacobian}
\label{sec:theory:approx}

\begin{figure}[t]
\centering
\begin{minipage}[t]{0.487\textwidth}
\centering
\includegraphics[width=\linewidth]{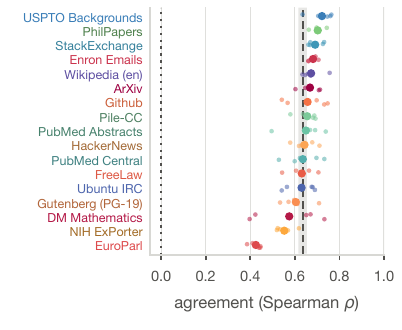}
\end{minipage}\hfill
\begin{minipage}[t]{0.487\textwidth}
\centering
\includegraphics[width=\linewidth]{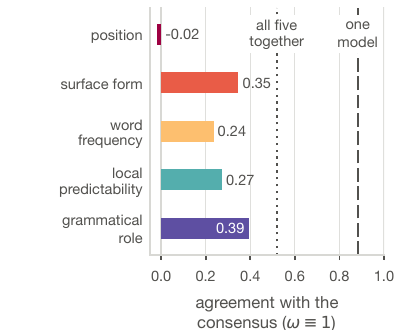}
\end{minipage}
\caption{\textbf{What the \tw{per-token marginal attention} tracks.} Spearman correlation per document. \textit{Left:} with the \iw{per-token marginal influence} of Qwen2.5-0.5B over 85 documents. \textit{Right:} word properties regressing the unweighted ($\omega \equiv 1$) consensus profile over 550 documents.}
\label{fig:jacobian_predictors}
\end{figure}

\textbf{Empirical evidence.} We compare the \tw{per-token marginal attention} $\tokw(X)$ of Qwen2.5-0.5B, with the weights $\omega$ of \cref{sec:method}, to its \iw{per-token marginal influence} $\infl(X)$ on the same document, across 85 natural documents. To approximate the Jacobian we apply Hutchinson's estimator to a random subset of blocks (\cref{app:theory_measurements:estimators}). The two quantities are compared as in \cref{sec:census}: both are projected onto the characters, converted to ranks, residualised against 50 relative-position bins, and compared by Spearman correlation, one document at a time. \Cref{fig:jacobian_predictors} (left) shows that the \tw{per-token marginal attention} and the \iw{per-token marginal influence} agree across all domains (average Spearman $\rho$ of 0.635). Agreement is lowest in the EuroParl domain, which contains substantial non-English text. As a sanity check, agreement of the two for a pair of distinct documents is nearly zero (0.022). As shown in \cref{app:theory_measurements:estimators}, the choice of weights does influence the level of agreement: the simplest setting of uniform weights $\omega \equiv 1$ reduces the agreement to 0.534.

Across models, where Hutchinson's estimate is costly, we approximate the \iw{per-token marginal influence} by a scalar projection of each Jacobian block, the gradient of the log-probability of the observed next token (\cref{app:theory_measurements:estimators}). Over 16 models from 6 families, each model's influence map agrees with the consensus of the other models at 0.772, and the unweighted \tw{per-token marginal attention} at 0.861 (\cref{tab:jacobian_census_headline}). \textbf{The link does not need attention at all}: the \iw{per-token marginal influence} of Mamba \citep{gu2024mamba} and RWKV-4 \citep{peng-etal-2023-rwkv} models, which have no attention mechanism, agrees with the \tw{per-token marginal attention} of the census similarly to a transformer model (\cref{app:theory_measurements:attention_free}).

\textbf{Derivation.} We now motivate this approximation heuristically. For a single causal attention layer with the attention weights held fixed, head $h$ adds $\sum_{s \le t} \attnel{h}{s}{t}\, W_{OV}^{(h)} \mathbf{x}_s$ to the residual stream at position $t$, with $W_{OV}^{(h)} := W_O^{(h)} W_V^{(h)}$, so its part of the Jacobian block is $[J_h^{(s)}]_t = \attnel{h}{s}{t} W_{OV}^{(h)}$, and the block $[J^{(s)}]_t$ is the sum of these parts over heads, up to the identity at $s = t$. Applied to $\mathbf{x}_s$, the head's part has norm $\attnel{h}{s}{t}\, \lVert W_O^{(h)} v_s \rVert = \wt{h}{s}{t}\, \attnel{h}{s}{t}$, exactly the summand of \cref{defn:marginalAttention}. Summing over output positions and heads, up to normalisation,
\begin{equation*}
    \tokwel{s} \propto \sum_{t \ge s} \sum_{h} \bigl\lVert [J_h^{(s)}]_t\, \mathbf{x}_s \bigr\rVert,
    \qquad
    \inflel{s} \propto \sum_{t \ge s} \Bigl\lVert \sum_{h} [J_h^{(s)}]_t \Bigr\rVert_F,
\end{equation*}
The two differ only in that $\tokw$ sums the heads' norms along $\mathbf{x}_s$ where $\infl$ takes the Frobenius norm of their sum, so they agree when the heads' contributions do not systematically cancel and $\mathbf{x}_s$ is a generic direction of roughly constant norm. We check the first condition directly in \cref{app:multilayer:single}. Along the training of Pythia-410M, the norm of the head-summed block stays within 5\% of its value for orthogonal heads, and on four finished models it is at most 17\% below it (\cref{fig:head_cancellation_training,fig:head_cancellation_models}). With $\omega \equiv 1$, the first-order term drops each head's gain $\lVert W_{OV}^{(h)} \mathbf{x}_s \rVert$, which may explain lower agreement. \Cref{app:multilayer:single} gives the full argument, and \cref{app:multilayer:multi} extends it to several layers.

\takeaway{The \tw{per-token marginal attention} $\tokw(X)$ is an empirically validated approximation to the \iw{per-token marginal influence} $\infl(X)$: it is the first-order, per-head part of the Jacobian.}

\subsection{Jacobian identifiability}
\label{sec:theory:identifiability}

Having established that \tw{per-token marginal attention} $\tokw(X)$ is empirically related to the \iw{per-token marginal influence} $\infl(X)$, we now develop a model-agnostic identifiability theory for $\infl(X)$.

Our theory relies on three assumptions, stated formally in \cref{app:proofs:assumptions} with empirical diagnostics reported in \cref{app:theory_measurements}: \textit{i)} \emph{Jacobian smoothness} (\cref{assump:jacobianSmoothness}) bounds the curvature of the model by a constant $M$; we probe local sensitivity with Jacobian-vector products; \textit{ii)} \emph{Preceding-token dependence} (\cref{assump:preTokenDependence}) states that each prediction depends non-trivially on the token immediately before it; \textit{iii)} \emph{lag decay} (\cref{assump:lagDecay}) states that the dependence of output positions on increasingly ``further back'' input positions decays toward zero. Our claim is one of statistical near-identifiability \citep{nelsonID}, with the constants in the hypotheses giving quantitative control over the level of identifiability.

\begin{figure}[t]
\centering
\includegraphics[width=\linewidth]{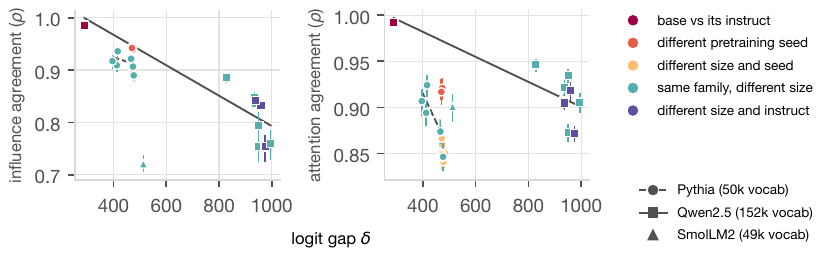}
\input{figures/main/theory_gap_caption.tex}
\label{fig:logit_gap}
\end{figure}

\begin{theorem}[Near-Identifiability of the Per-Token Marginal Influence, informal]
    \label{thm:jacobianControl}
    Let $f, f'$ be two language models satisfying \cref{assump:jacobianSmoothness,assump:preTokenDependence,assump:lagDecay}. If their next-token logits agree under small perturbations of any single input position,
    \begin{equation*}
        \left\lVert f_t(X^{(s) + r\mathbf{u}}) - f_t'(X^{(s) + r\mathbf{u}}) \right\rVert \leq \delta \qquad \text{for all } s \leq t, \ 0 \leq r < R, \ \lVert \mathbf{u} \rVert = 1,
    \end{equation*}
    then their \iw{per-token marginal influences} agree,
    \begin{equation*}
        \lVert \infl(X) - \infl'(X) \rVert_1 \leq \epsilon,
    \end{equation*}
    where $\epsilon$ depends on neither $n$ nor $d$, and $\epsilon \to 0$ as $M, \delta \to 0$.
\end{theorem}

Training produces a small logit gap $\delta$: two models fit to the same next-token objective approach the text's conditional distribution as their loss falls. Over a training run, cross-family agreement rises as the loss falls and is near zero until it beats a bigram baseline (\cref{fig:emergence,app:theory_measurements:emergence}).

\textbf{Experimental evidence.} We take the 16 models of \cref{sec:theory:approx} and form 26 tokenizer-sharing pairs: 15 pythia pairs, 10 qwen2.5 pairs and one SmolLM2 pair, evaluated on 100 documents. The logit gap $\delta$ of a pair is the norm of the difference of the two models' centred next-token logits, averaged over positions and documents. Agreement (\cref{sec:census}) is computed on each tokenizer group's character axis and applied both to the \iw{per-token marginal influence}, in the row-gradient approximation of \cref{sec:theory:approx}, and to the unweighted ($\omega \equiv 1$) \tw{per-token marginal attention}. Within a tokenizer group, as the logit gap grows, \iw{per-token marginal influence} agreement falls, which is consistent with \cref{thm:jacobianControl}. Validating our empirical argument, \tw{per-token marginal attention} agreement falls as well.

\takeaway{Sufficiently smooth models with similar next-token logits for a given input have their \iw{per-token marginal influences} guaranteed to be close, under mild assumptions on the dependence patterns in the Jacobian (\cref{thm:jacobianControl}). Empirically, a smaller logit gap $\delta$ implies higher agreement on both \iw{per-token marginal influence} and \tw{per-token marginal attention} (\cref{fig:logit_gap}).}

\section{Application: KV cache eviction}
\label{sec:application}

KV cache eviction bounds the memory of a long context by keeping only part of the stored keys and values (\cref{sec:related:eviction}). Eviction decides \textit{how many} entries each head keeps and \textit{which} ones, so we write the retention score of an entry as a \hw{per-head} budget times a per-entry score:
\begin{equation}
    \label{eq:retention}
    S_{h,s}(X) \;=\; \hw{\beta_h} \cdot \gamma_{h,s}(X),
\end{equation}
where $\hw{\beta_h} \ge 0$ is the budget of head $h$, the fraction of its entries it keeps, and $\gamma_{h,s}(X) \ge 0$ is the score of entry $(h,s)$ in document $X$. Head $h$ keeps its $\hw{\beta_h} n$ highest-scoring entries. Budgets are set per key-value head, and in this section $H$ is the number of key-value heads. We ask whether a static per-model budget can replace document-specific ones, and answer with a budget read from the \hw{per-head marginal attention} (\cref{sec:application:budget}). To obtain an eviction policy, we pair it with Expected Attention as the score (\cref{sec:application:results}).

\subsection{Head budget from the \texorpdfstring{\hw{per-head marginal attention}}{per-head marginal attention}}
\label{sec:application:budget}

\textbf{\headmass{}} (HM) is the \hw{per-head marginal attention} of \cref{defn:perHead} with three choices taken from KVzip, a state-of-the-art eviction method that is expensive at run time, since it pays a second target forward per document. First, \headmass{} is read on the \textbf{reconstruction queries}, the positions $t$ of a second pass in which the model repeats the context, since eviction must keep what later queries attend to. Second, the per-entry score of KVzip+ \citep{jegou2026kvzapfastadaptivefaithful} replaces the mean over queries of \cref{defn:marginalAttention}: an entry scores the largest value-weighted attention it receives from those queries, normalised by the norm of the attention input, since an entry that any one later query needs must be kept (exact form in \cref{eq:headmass_share}, \cref{app:kv_details:budget}). Third, \headmass{} applies KVzip's global cut, which keeps the fraction $B$ of all entries, so head $h$ counts the fraction $m_h(X)$ of its own entries the cut keeps.

The head share $h$ is $p_h(X) = m_h(X) / \sum_{j} m_j(X)$. We average it over a set $\mathcal{D}_{\text{pre}}$ of 200 pretraining documents disjoint from benchmarks (\cref{app:kv_details:budget}), and the budget spends $B$ with this proportion:
\begin{equation}
    \label{eq:budget}
    \hw{\beta_h} = H B \, \headwel{h},
    \qquad
    \headwel{h} = \mathbb{E}_{X \sim \mathcal{D}_{\text{pre}}}\bigl[p_h(X)\bigr],
\end{equation}
clipped to a floor of $0.01$ and a cap of $1$. The budgets are then rescaled so that the heads spend exactly $B$ on average. The \hw{per-head} profile is a property of the model (\cref{sec:census}) and stays so on the reconstruction queries (\cref{tab:kv_budget_stability}), so the budget is computed once per model.

\subsection{Results}
\label{sec:application:results}

\begin{table}[t]
\centering
\caption{\textbf{Eviction scores.} RULER 4096 score over its 13 tasks and documents at 4$\times$, 8$\times$ and 16$\times$ compression. \textit{Cost:} $\triangle$ offline work per target, $\bullet$ online work per document (\cref{tab:kv_cost}). \textit{Shade:} rank within a column, darker is better; \textit{Bold:} best method; the uncompressed row is not ranked. \textit{Italics:} our run, where no budget curve or surrogate is published for the target.}

\label{tab:kv_ruler_4096}
\begingroup
\footnotesize
\setlength{\tabcolsep}{1.6pt}
\ifdefined\kvshadecolw\else\newlength{\kvshadecolw}\fi
\settowidth{\kvshadecolw}{\footnotesize\bfseries Llama-3.1-8B}
\addtolength{\kvshadecolw}{-4\tabcolsep}
\divide\kvshadecolw by 3
\ifdefined\kvshadecolwfl\else\newlength{\kvshadecolwfl}\fi
\settowidth{\kvshadecolwfl}{\footnotesize\bfseries 88.4}
\ifdim\kvshadecolw<\kvshadecolwfl\setlength{\kvshadecolw}{\kvshadecolwfl}\fi
\begin{tabular}{>{\raggedright\arraybackslash}m{0.9in}cc>{\centering\arraybackslash}m{\kvshadecolw}>{\centering\arraybackslash}m{\kvshadecolw}>{\centering\arraybackslash}m{\kvshadecolw}>{\centering\arraybackslash}m{\kvshadecolw}>{\centering\arraybackslash}m{\kvshadecolw}>{\centering\arraybackslash}m{\kvshadecolw}>{\centering\arraybackslash}m{\kvshadecolw}>{\centering\arraybackslash}m{\kvshadecolw}>{\centering\arraybackslash}m{\kvshadecolw}>{\centering\arraybackslash}m{\kvshadecolw}>{\centering\arraybackslash}m{\kvshadecolw}>{\centering\arraybackslash}m{\kvshadecolw}>{\centering\arraybackslash}m{\kvshadecolw}>{\centering\arraybackslash}m{\kvshadecolw}>{\centering\arraybackslash}m{\kvshadecolw}}
\toprule
 & \multicolumn{2}{c}{\textbf{cost}} & \multicolumn{3}{c}{\textbf{Qwen3-8B}} & \multicolumn{3}{c}{\textbf{Llama-3.1-8B}} & \multicolumn{3}{c}{\textbf{phi-4}} & \multicolumn{3}{c}{\textbf{Qwen3-32B}} & \multicolumn{3}{c}{\textbf{Mistral-7B}} \\
\cmidrule(lr){2-3}\cmidrule(lr){4-6}\cmidrule(lr){7-9}\cmidrule(lr){10-12}\cmidrule(lr){13-15}\cmidrule(lr){16-18}
 & \scriptsize offline & \scriptsize online & 4$\times$ & 8$\times$ & 16$\times$ & 4$\times$ & 8$\times$ & 16$\times$ & 4$\times$ & 8$\times$ & 16$\times$ & 4$\times$ & 8$\times$ & 16$\times$ & 4$\times$ & 8$\times$ & 16$\times$ \\
\midrule
AdaKV + EA &  &  & \cellcolor{kvshade!12}88.4 & \cellcolor{kvshade!12}72.8 & \cellcolor{kvshade!12}46.5 & \cellcolor{kvshade!12}75.7 & \cellcolor{kvshade!12}45.3 & \cellcolor{kvshade!12}15.2 & \cellcolor{kvshade!12}68.7 & \cellcolor{kvshade!12}44.0 & \cellcolor{kvshade!12}13.8 & \cellcolor{kvshade!12}81.9 & \cellcolor{kvshade!12}63.6 & \cellcolor{kvshade!12}36.4 & \cellcolor{kvshade!12}67.2 & \cellcolor{kvshade!12}34.0 & \cellcolor{kvshade!12}19.9 \\
AdaKV + KVzap & $\triangle\triangle$ &  & \cellcolor{kvshade!47}94.0 & \cellcolor{kvshade!47}89.2 & \cellcolor{kvshade!30}49.9 & \cellcolor{kvshade!30}94.5 & \cellcolor{kvshade!30}89.3 & \cellcolor{kvshade!30}63.1 & \cellcolor{kvshade!30}\textit{87.8} & \cellcolor{kvshade!30}\textit{77.0} & \cellcolor{kvshade!30}\textit{51.6} & \cellcolor{kvshade!82}95.4 & \cellcolor{kvshade!47}90.0 & \cellcolor{kvshade!47}50.3 & \cellcolor{kvshade!30}\textit{88.3} & \cellcolor{kvshade!30}\textit{70.2} & \cellcolor{kvshade!30}\textit{42.1} \\
KVzip &  & $\bullet\bullet$ & \cellcolor{kvshade!82}95.0 & \cellcolor{kvshade!100}\textbf{92.3} & \cellcolor{kvshade!65}59.4 & \cellcolor{kvshade!82}95.3 & \cellcolor{kvshade!82}92.8 & \cellcolor{kvshade!100}\textbf{74.2} & \cellcolor{kvshade!65}91.5 & \cellcolor{kvshade!82}89.6 & \cellcolor{kvshade!65}68.8 & \cellcolor{kvshade!100}\textbf{95.6} & \cellcolor{kvshade!100}\textbf{95.6} & \cellcolor{kvshade!100}\textbf{91.8} & \cellcolor{kvshade!65}91.2 & \cellcolor{kvshade!100}\textbf{89.2} & \cellcolor{kvshade!100}\textbf{72.2} \\
LU-KV + EA & $\triangle$ &  & \cellcolor{kvshade!30}\textit{90.6} & \cellcolor{kvshade!30}\textit{75.2} & \cellcolor{kvshade!47}\textit{58.4} & \cellcolor{kvshade!65}95.1 & \cellcolor{kvshade!47}91.8 & \cellcolor{kvshade!65}67.0 & \cellcolor{kvshade!100}\textit{\textbf{92.4}} & \cellcolor{kvshade!65}\textit{87.9} & \cellcolor{kvshade!82}\textit{69.5} & \cellcolor{kvshade!30}\textit{90.2} & \cellcolor{kvshade!30}\textit{72.6} & \cellcolor{kvshade!30}\textit{38.2} & \cellcolor{kvshade!82}91.2 & \cellcolor{kvshade!65}85.1 & \cellcolor{kvshade!47}61.7 \\
\hw{HM} + EA & $\triangle$ &  & \cellcolor{kvshade!65}95.0 & \cellcolor{kvshade!82}91.8 & \cellcolor{kvshade!100}\textbf{73.3} & \cellcolor{kvshade!47}95.1 & \cellcolor{kvshade!65}92.3 & \cellcolor{kvshade!47}66.3 & \cellcolor{kvshade!47}91.5 & \cellcolor{kvshade!47}87.8 & \cellcolor{kvshade!47}68.5 & \cellcolor{kvshade!47}95.2 & \cellcolor{kvshade!65}93.2 & \cellcolor{kvshade!65}79.1 & \cellcolor{kvshade!47}90.5 & \cellcolor{kvshade!47}84.3 & \cellcolor{kvshade!65}65.3 \\
\hw{HM} + Guide & $\triangle$ & $\bullet$ & \cellcolor{kvshade!100}\textbf{95.1} & \cellcolor{kvshade!65}91.6 & \cellcolor{kvshade!82}71.3 & \cellcolor{kvshade!100}\textbf{95.3} & \cellcolor{kvshade!100}\textbf{93.5} & \cellcolor{kvshade!82}72.3 & \cellcolor{kvshade!82}92.3 & \cellcolor{kvshade!100}\textbf{91.2} & \cellcolor{kvshade!100}\textbf{73.9} & \cellcolor{kvshade!65}95.4 & \cellcolor{kvshade!82}94.1 & \cellcolor{kvshade!82}79.2 & \cellcolor{kvshade!100}\textbf{91.3} & \cellcolor{kvshade!82}86.7 & \cellcolor{kvshade!82}71.6 \\
\midrule
Uncompressed &  &  & \multicolumn{3}{c}{95.4} & \multicolumn{3}{c}{95.7} & \multicolumn{3}{c}{91.0} & \multicolumn{3}{c}{95.7} & \multicolumn{3}{c}{92.2} \\
\bottomrule
\end{tabular}
\endgroup

\end{table}

\paragraph{Experimental setting.}
We evaluate \cref{eq:retention} on Qwen3-8B, Qwen3-32B, Llama-3.1-8B-Instruct, phi-4 and Mistral-7B-Instruct-v0.3 \citep{yang2025qwen3technicalreport,grattafiori2024llama3herdmodels,abdin2024phi4technicalreport,jiang2023mistral7b}. Each method pairs a budget with a score. AdaKV~+~EA \citep{NEURIPS2025_a40ff56d} splits the budget inside each layer per document and scores with Expected Attention (EA) \citep{devoto2025expected}, a training-free scorer. KVzip \citep{NEURIPS2025_f4eaa4b8} takes both the budget and the score from a second target forward per document. AdaKV~+~KVzap \citep{jegou2026kvzapfastadaptivefaithful} keeps the AdaKV budget and scores with a KVzip surrogate trained per target. LU-KV~+~EA \citep{tang2026predictingfutureutilityglobal} takes its budget from an offline per-head table calibrated per target and scorer, and scores with EA. \hw{HM}~+~EA takes our \headmass{} budget and scores with EA. \hw{HM}~+~Guide keeps the \headmass{} budget and tilts EA, head by head, by a linear map fitted once per target (\cref{app:kv_details:guide}). The map predicts the KVzip+ score from the marginal attention of Qwen2.5-0.5B-Instruct, a model about fifteen times smaller than the smallest target. It probes whether a smaller model's attention carries the eviction signal and is not meant as a practical method. We report RULER \citep{hsieh2024ruler} (13 tasks, 6500 documents) at context 4096 and compression $4\times$, $8\times$ and $16\times$, the factor by which the cache shrinks. Scores at context 8192 ($8\times$ and $16\times$) and on LongBench \citep{bai-etal-2024-longbench} are in \cref{tab:kv_ruler_8192,tab:kv_longbench}, per-task scores in \cref{app:kv_details:per_task}, and costs in \cref{tab:kv_cost}.

\paragraph{Result analysis.}
\textit{Budget.} \Cref{tab:kv_ruler_4096} shows that the \hw{HM} budget accounts for most of the gap from standard EA to KVzip. In fact, with the same scorer, \hw{HM}~+~EA considerably improves on its baseline on every target and setting, closing most of the gap to KVzip without the double-forward on the target. \hw{HM}~+~EA is also ahead of AdaKV~+~KVzap in all but one cell, and its budget is built offline in minutes and read at inference as a fixed table, with no representation access, where a KVzap surrogate trains for hours (\cref{tab:kv_cost}) and needs light online computation. Even though it has extra scorer calibration, LU-KV~+~EA is level with \hw{HM}~+~EA on phi-4 and Llama-3.1-8B, close on Mistral-7B, and far behind on both Qwen3 targets. Averaging over pretraining text loses little: the \hw{HM} budget agrees closely, head for head, with the keep rates KVzip+ realises on held-out documents (\cref{app:kv_details:budget_maps}). The query set is crucial: set by the \textbf{reading queries} (i.e., all positions), a proportional budget (\cref{eq:budget_write}) consistently loses to the baseline AdaKV~+~EA with no fixed budget (\cref{tab:kv_write_equivalence}), showing a profile can be stable across documents and still be the wrong quantity for eviction. The cut matters less (\cref{app:kv_details:write}).
\textit{Score.} Tilting EA by the guide's map raises \hw{HM}~+~EA in all but two columns, so a smaller model's attention can carry part of the eviction signal.

\takeaway{Which heads matter for eviction is a property of the model: a \hw{per-head marginal attention} profile computed once recovers most of the gain of per-document signals, and a score tilted by a smaller model can further improve it.}

\section{Conclusions}
\label{sec:conclusion}

We studied attention in language models through its marginals and found that one structure per document carries two different signals. Reduced over heads, the \tw{per-token marginal attention} is a property of the text: models that differ in lab, data, size, tokenizer and architecture agree on which input tokens are attended. Reduced over tokens, the \hw{per-head marginal attention} is a property of the model: it is stable across documents and private to each model.
We showed that the \tw{per-token profile} is shared because it tracks a Jacobian statistic that similar predictors must share: smooth models with close next-token distributions have close input-output Jacobians, with marginal attention approximating its the blockwise norms. 
The only data assumption is that its discrete conditional distributions admit a smooth approximation good enough for independently trained models to agree on the next token. 
To our knowledge this is the first formal result of the kind discussed by the Platonic Representation Hypothesis \citep{pmlr-v235-huh24a}, but it does not imply semantic convergence~\citep{nelsonID}: the shared profile separates documents without aligning strongly with classical syntactic structure.
The \hw{per-head profile} is stable enough to be computed once per model, offline, on pretraining text, and used as a budget for KV cache eviction. At inference only a per-token score remains, and since that score is shared across models it can be supplied by a much smaller one. The resulting method is competitive with those that recompute the budget on every document or train it per target.

\textbf{Limitations.}
\label{sec:limitations} We give empirical evidence that the \tw{per-token marginal attention} is shared across models and that the \hw{per-head marginal attention} is stable across documents, but our theory covers only the \iw{per-token marginal influence}. The link from the marginal influence to the marginal attention is empirical, and a study of the assumptions under which it holds would be of significant interest. The census covers mostly English text and reads at most 2048 tokens per document, the context limit of several of its older models: a claim that the profile belongs to the text and not to the model needs models that differ as much as possible, so we kept them at the cost of document length. \hw{HM}~+~Guide needs one extra forward pass of a small model for every document, which standard serving stacks do not support, so we use it as a probe of whether a smaller model's marginal attention carries the eviction signal, not as a practical method.

\section*{Acknowledgments}
This research was supported by the ISTA Responsible AI Program, made possible through the generous support of Garrett Camp and the Camp Foundation. WN was supported by the Chan Zuckerberg Initiative
(CZI) through the AI Residency Program. We thank CZI for
the opportunity to participate in this program.

\section*{AI use statement}

We used AI tools (Claude and Claude Code, running Claude Fable 5.1 and Claude Opus 5), during this work. The authors set the questions, the claims and the experiments, reviewed every contribution of the tools before it was kept, and take responsibility for the final content.

\paragraph{Experiments.} The authors wrote the core library by hand (with autocompletion). AI tools were used to write scripts and manage experiments. Results and code were reviewed by an author.

\paragraph{Writing and presentation.} AI tools were used to integrate raw results, perform bulk edits and layout changes, and edit the paper. \Cref{fig:teaser} was drawn by an AI tool from a sketch by the authors.

\paragraph{Theory.} The definitions, hypotheses, theorems and proofs were proposed by the authors. AI tools were used to find gaps in the argument and improve the exposition. Every definition, statement and proof was checked by the authors.

\paragraph{Model citations.} Given the large number of models used in this work, AI tools were used to scrape the citations for each model from the corresponding HuggingFace repository. These citations were checked by an author.

\section*{Reproducibility statement}

The code is available at \url{https://github.com/Flegyas/marginal-attention}. One library implements the estimator (the marginal attention hooks, the projection onto the character axis, ranking, positional residualization and the decomposition of \cref{sec:method}); each experiment using the library is self-contained. Figures and tables are end-to-end reproducible in notebooks included in the repository.

All 67 census models are public checkpoints, listed with their sources in \cref{app:census_details:models}. The code release includes a script to build the corpus used in the experiments. The KV eviction experiments use the \texttt{kvpress} library \citep{devoto2025expected} and its RULER and LongBench harnesses unchanged, with our budget and scorer added as one more press. \Cref{app:agreement} reports the controls, the seed ceiling and the sensitivity of the headline to our design choices; \cref{app:theory_measurements} reports empirical diagnostics of the assumptions of \cref{sec:theory}, and \cref{app:proofs} gives complete theorem statements and proofs with their assumptions stated formally.

\bibliography{iclr2027_conference,models}
\bibliographystyle{iclr2027_conference}

\appendix
\crefalias{section}{appendix}
\crefalias{subsection}{subappendix}
\crefalias{subsubsection}{subsubappendix}

\section{Census details}
\label{app:census_details}

\subsection{Data}
\label[appendix]{app:census_details:data}

The census corpus, \emph{mix1000}, draws 50 documents from each of 20 domains. Most domains are subsets of the Pile \citep{gao2020pile}, taken from its release without the copyrighted subsets (\texttt{monology/pile-uncopyrighted}). The other three are generated text with little linguistic structure: one short sentence repeated, a random string over the four DNA letters, and rows of random ASCII-art symbols. The main text uses only the natural documents.

\begin{figure}[t]
\centering
\includegraphics[width=0.5\linewidth]{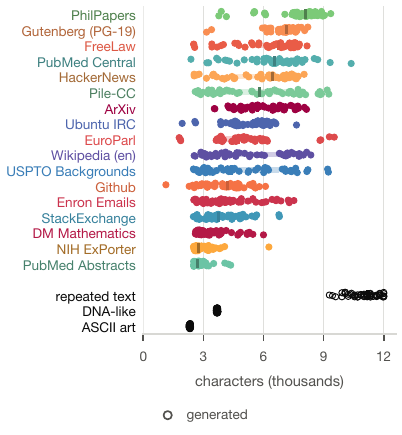}
\input{figures/main/census_lengths_caption.tex}
\label{fig:census_lengths}
\end{figure}

Pile documents shorter than 2500 characters are skipped, and the documents of each domain are drawn by seeded reservoir sampling over one pass of the stream. Each document contributes one window of 12{,}000 characters, taken from a random position inside it. Extraction stops at 2048 tokens per model, and each document is scored only on the characters that every model covers. \Cref{fig:census_lengths} shows the length of every document by domain.

A second corpus, \emph{mix10k}, follows the same recipe with ten times as many documents per domain. It samples Pile rows that the first corpus never reads, and its generated domains use a different seed, so the two corpora are practically disjoint. Ubuntu IRC is capped lower than the other domains, because the Pile holds too few of its documents above the length floor.

\begin{figure}[t]
\centering
\includegraphics[width=\linewidth]{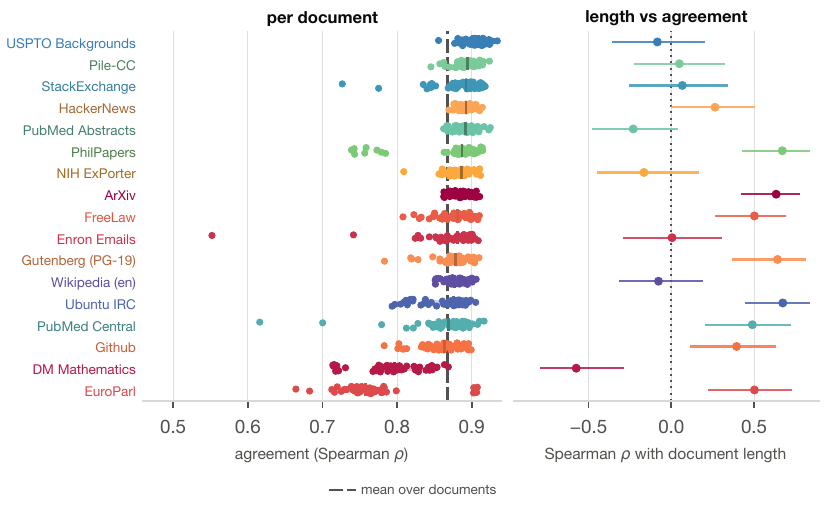}
\input{figures/main/census_documents_caption.tex}
\label{fig:census_documents}
\end{figure}

\begin{figure}[t]
\centering
\includegraphics[width=0.5\linewidth]{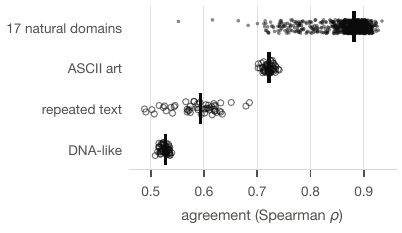}
\input{figures/main/synthetic_domains_caption.tex}
\label{fig:synthetic_domains}
\end{figure}

\Cref{tab:census_arms} applies the estimator of \cref{sec:census} to each corpus, on all of its documents, on the natural ones and on the generated ones. Agreement is lower on generated text; on the larger corpus, scored with a twelve-model subset of the roster, it stays high, so the headline does not depend on one draw of documents. \Cref{fig:census_documents} shows the agreement of each natural document by domain, and how it relates to document length within each domain. \Cref{fig:synthetic_domains} shows the per-domain agreement on the three generated domains.

\begin{table}[htbp]
\centering
\caption{\textbf{\tw{Per-token} agreement on every document set.} Each row applies the same estimator to the value-weighted \tw{per-token marginal attention} on one set of documents: \emph{mix1000} (20 kinds of text), its natural and generated subsets, and \emph{mix10k}. $\pm$ is half the 95\% bootstrap interval over documents.}

\label{tab:census_arms}
\begingroup
\footnotesize
\setlength{\tabcolsep}{3pt}
\begin{tabular}{l r r r r}
\toprule
document set & models & documents & consensus of the other models & between two models \\
\midrule
mix1000, all domains & 67 & 1000 & 0.8443 $\pm$ 0.0042 & 0.7253 $\pm$ 0.0065 \\
mix1000, natural domains & 67 & 850 & 0.8673 $\pm$ 0.0032 & 0.7607 $\pm$ 0.0051 \\
mix1000, generated domains & 67 & 150 & 0.6114 $\pm$ 0.0136 & 0.4093 $\pm$ 0.0162 \\
\midrule
mix10k, all domains & 12 & 9792 & 0.7977 $\pm$ 0.0021 & 0.6787 $\pm$ 0.0028 \\
mix10k, natural domains & 12 & 8292 & 0.8357 $\pm$ 0.0012 & 0.7277 $\pm$ 0.0018 \\
mix10k, generated domains & 12 & 1500 & 0.4990 $\pm$ 0.0064 & 0.3362 $\pm$ 0.0060 \\
\bottomrule
\end{tabular}
\endgroup

\end{table}

\subsection{Models}
\label[appendix]{app:census_details:models}

\Cref{tab:model_roster} lists every model of the census, with its family, architecture and size, and \cref{tab:census_main} gives the census summary.

\begin{table}[htbp]
\centering
\caption{\textbf{Per-token census summary.} 67 models, 850 natural documents. \textit{Upper:} agreement with the other 66 models and between two models, with the different-document control and the pythia-410m seed ceiling. \textit{Lower:} agreement between two models by pair class. On the first row, $\pm$ is half the 95\% bootstrap interval over documents.}

\label{tab:census_main}
\begingroup
\footnotesize
\setlength{\tabcolsep}{3pt}
\begin{tabular}{clrrrr}
\toprule
 &  & \emph{vs the other 66} & \emph{two models} &  &  \\
\cmidrule(lr){3-4}
\multirow{3}{*}{\rotatebox[origin=c]{90}{\shortstack{whole\\census}}} & agreement & 0.867 $\pm$ 0.003 & 0.761 $\pm$ 0.005 &  &  \\
 & different document (control) & 0.015 & 0.014 &  &  \\
 & 10 pythia-410m seeds (ceiling) & 0.942 & 0.900 &  &  \\
\cmidrule(lr){1-6}
 &  & \emph{pairs} & \emph{mean} & \emph{median} & \emph{5th to 95th} \\
\cmidrule(lr){3-6}
\multirow{6}{*}{\rotatebox[origin=c]{90}{\shortstack{pair\\class}}} & base vs its instruct & 12 & 0.968 & 0.976 & [0.924, 0.993] \\
 & vision vs its text model & 3 & 0.941 & 0.944 & [0.928, 0.951] \\
 & same family, different size & 22 & 0.889 & 0.904 & [0.807, 0.945] \\
 & different pretraining seed & 225 & 0.821 & 0.849 & [0.628, 0.933] \\
 & same architecture, different lab & 28 & 0.788 & 0.784 & [0.693, 0.896] \\
 & different lab and data & 85 & 0.755 & 0.777 & [0.547, 0.868] \\
\bottomrule
\end{tabular}
\endgroup

\end{table}

\begin{table}[htbp]
\centering
\caption{\textbf{Every model used.} Repository id, parameter count in billions (B) and the report the model card names (ref.); alphabetical by repository, down the left block and then down the right. 69 models: 67 form both the \tw{per-token} census and the \hw{per-head} analyses, and Qwen3-32B and Mistral-7B-Instruct-v0.3 are used only as eviction targets. Eviction targets: Qwen3-8B, Llama-3.1-8B-Instruct, phi-4, Qwen3-32B, Mistral-7B-Instruct-v0.3; guide: Qwen2.5-0.5B-Instruct. The Pythia pretraining-seed pool (10 seeds at each of 5 sizes) is not listed.}

\label{tab:model_roster}
\begingroup
\scriptsize
\setlength{\tabcolsep}{2pt}
\begin{tabular}{>{\raggedright\arraybackslash}p{92pt} r >{\raggedright\arraybackslash}p{72pt} >{\raggedright\arraybackslash}p{92pt} r >{\raggedright\arraybackslash}p{72pt}}
\toprule
model & B & ref. & model & B & ref. \\
\midrule
\texttt{01-\allowbreak ai/\allowbreak Yi-\allowbreak 6B} & 6.1 & \citep{ai2024yiopenfoundationmodels} & \texttt{meta-\allowbreak llama/\allowbreak Llama-\allowbreak 3.1-\allowbreak 8B} & 8.0 & \citep{grattafiori2024llama3herdmodels} \\
\texttt{01-\allowbreak ai/\allowbreak Yi-\allowbreak 6B-\allowbreak Chat} & 6.1 & \citep{ai2024yiopenfoundationmodels} & \texttt{meta-\allowbreak llama/\allowbreak Llama-\allowbreak 3.1-\allowbreak 8B-\allowbreak Instruct} & 8.0 & \citep{grattafiori2024llama3herdmodels} \\
\texttt{01-\allowbreak ai/\allowbreak Yi-\allowbreak 9B} & 8.8 & \citep{ai2024yiopenfoundationmodels} & \texttt{microsoft/\allowbreak phi-\allowbreak 4} & 14.7 & \citep{abdin2024phi4technicalreport} \\
\texttt{allenai/\allowbreak OLMo-\allowbreak 2-\allowbreak 0425-\allowbreak 1B} & 1.5 & \citep{walsh2025olmo2furious} & \texttt{mistralai/\allowbreak Mistral-\allowbreak 7B-\allowbreak Instruct-\allowbreak v0.1} & 7.2 & \citep{jiang2023mistral7b} \\
\texttt{allenai/\allowbreak OLMo-\allowbreak 2-\allowbreak 0425-\allowbreak 1B-\allowbreak Instruct} & 1.5 & \citep{walsh2025olmo2furious} & \texttt{mistralai/\allowbreak Mistral-\allowbreak 7B-\allowbreak Instruct-\allowbreak v0.2} & 7.2 & \citep{jiang2023mistral7b} \\
\texttt{allenai/\allowbreak OLMoE-\allowbreak 1B-\allowbreak 7B-\allowbreak 0924} & 6.9 & \citep{muennighoff2025olmoe} & \texttt{mistralai/\allowbreak Mistral-\allowbreak 7B-\allowbreak Instruct-\allowbreak v0.3} & 7.2 & \citep{jiang2023mistral7b} \\
\texttt{bigcode/\allowbreak starcoder2-\allowbreak 3b} & 3.0 & \citep{lozhkov2024starcoder2stackv2} & \texttt{mistralai/\allowbreak Mistral-\allowbreak 7B-\allowbreak v0.1} & 7.2 & \citep{jiang2023mistral7b} \\
\texttt{bigscience/\allowbreak bloom-\allowbreak 1b7} & 1.7 & \citep{workshop2023bloom} & \texttt{mistralai/\allowbreak Mixtral-\allowbreak 8x7B-\allowbreak v0.1} & 46.7 & \citep{jiang2024mixtralexperts} \\
\texttt{CohereLabs/\allowbreak aya-\allowbreak expanse-\allowbreak 8b} & 8.0 & \citep{dang2024ayaexpansecombiningresearch} & \texttt{nvidia/\allowbreak Mistral-\allowbreak NeMo-\allowbreak Minitron-\allowbreak 8B-\allowbreak Base} & 8.4 & \citep{sreenivas2024llmpruningdistillationpractice,muralidharan2024compactlanguagemodelspruning} \\
\texttt{deepseek-\allowbreak ai/\allowbreak deepseek-\allowbreak coder-\allowbreak 1.3b-\allowbreak base} & 1.3 & \citep{guo2024deepseekcoderlargelanguagemodel} & \texttt{openai/\allowbreak gpt-\allowbreak oss-\allowbreak 120b} & 116.8 & \citep{openai2025gptoss} \\
\texttt{deepseek-\allowbreak ai/\allowbreak DeepSeek-\allowbreak V2-\allowbreak Lite} & 15.7 & \citep{deepseekai2024deepseekv2} & \texttt{openai/\allowbreak gpt-\allowbreak oss-\allowbreak 20b} & 21.5 & \citep{openai2025gptoss} \\
\texttt{EleutherAI/\allowbreak gpt-\allowbreak j-\allowbreak 6b} & 6.1 & \citep{wang2021gptj} & \texttt{openbmb/\allowbreak MiniCPM3-\allowbreak 4B} & 4.1 & \citep{hu2024minicpmunveilingpotentialsmall} \\
\texttt{EleutherAI/\allowbreak gpt-\allowbreak neo-\allowbreak 1.3B} & 1.4 & \citep{black2021gptneo} & \texttt{OpenGVLab/\allowbreak InternVL3-\allowbreak 8B-\allowbreak hf} & 7.9 & \citep{zhu2025internvl3} \\
\texttt{EleutherAI/\allowbreak pythia-\allowbreak 1.4b} & 1.5 & \citep{pmlr-v202-biderman23a} & \texttt{Qwen/\allowbreak Qwen2.5-\allowbreak 0.5B} & 0.5 & \citep{qwen2025qwen25technicalreport} \\
\texttt{EleutherAI/\allowbreak pythia-\allowbreak 2.8b} & 2.9 & \citep{pmlr-v202-biderman23a} & \texttt{Qwen/\allowbreak Qwen2.5-\allowbreak 0.5B-\allowbreak Instruct} & 0.5 & \citep{qwen2025qwen25technicalreport} \\
\texttt{EleutherAI/\allowbreak pythia-\allowbreak 410m} & 0.5 & \citep{pmlr-v202-biderman23a} & \texttt{Qwen/\allowbreak Qwen2.5-\allowbreak 1.5B} & 1.5 & \citep{qwen2025qwen25technicalreport} \\
\texttt{EleutherAI/\allowbreak pythia-\allowbreak 6.9b} & 7.0 & \citep{pmlr-v202-biderman23a} & \texttt{Qwen/\allowbreak Qwen2.5-\allowbreak 1.5B-\allowbreak Instruct} & 1.5 & \citep{qwen2025qwen25technicalreport} \\
\texttt{facebook/\allowbreak opt-\allowbreak 1.3b} & 1.3 & \citep{zhang2022optopenpretrainedtransformer} & \texttt{Qwen/\allowbreak Qwen2.5-\allowbreak 32B} & 32.8 & \citep{qwen2025qwen25technicalreport} \\
\texttt{facebook/\allowbreak opt-\allowbreak 6.7b} & 6.7 & \citep{zhang2022optopenpretrainedtransformer} & \texttt{Qwen/\allowbreak Qwen2.5-\allowbreak 32B-\allowbreak Instruct} & 32.8 & \citep{qwen2025qwen25technicalreport} \\
\texttt{google/\allowbreak gemma-\allowbreak 2-\allowbreak 2b} & 2.6 & \citep{gemmateam2024gemma2} & \texttt{Qwen/\allowbreak Qwen2.5-\allowbreak 3B} & 3.1 & \citep{qwen2025qwen25technicalreport} \\
\texttt{google/\allowbreak gemma-\allowbreak 2-\allowbreak 2b-\allowbreak it} & 2.6 & \citep{gemmateam2024gemma2} & \texttt{Qwen/\allowbreak Qwen2.5-\allowbreak 7B} & 7.6 & \citep{qwen2025qwen25technicalreport} \\
\texttt{google/\allowbreak gemma-\allowbreak 2-\allowbreak 9b} & 9.2 & \citep{gemmateam2024gemma2} & \texttt{Qwen/\allowbreak Qwen2.5-\allowbreak 7B-\allowbreak Instruct} & 7.6 & \citep{qwen2025qwen25technicalreport} \\
\texttt{google/\allowbreak gemma-\allowbreak 2-\allowbreak 9b-\allowbreak it} & 9.2 & \citep{gemmateam2024gemma2} & \texttt{Qwen/\allowbreak Qwen2.5-\allowbreak Coder-\allowbreak 7B} & 7.6 & \citep{hui2024qwen25codertechnicalreport} \\
\texttt{google/\allowbreak gemma-\allowbreak 3-\allowbreak 1b-\allowbreak it} & 1.0 & \citep{gemmateam2025gemma3} & \texttt{Qwen/\allowbreak Qwen2.5-\allowbreak VL-\allowbreak 7B-\allowbreak Instruct} & 8.3 & \citep{bai2025qwen25vltechnicalreport} \\
\texttt{google/\allowbreak gemma-\allowbreak 3-\allowbreak 270m} & 0.3 & \citep{gemmateam2025gemma3} & \texttt{Qwen/\allowbreak Qwen3-\allowbreak 32B} & 32.8 & \citep{yang2025qwen3technicalreport} \\
\texttt{HuggingFaceTB/\allowbreak SmolLM2-\allowbreak 1.7B} & 1.7 & \citep{allal2025smollm2smolgoesbig} & \texttt{Qwen/\allowbreak Qwen3-\allowbreak 8B} & 8.2 & \citep{yang2025qwen3technicalreport} \\
\texttt{HuggingFaceTB/\allowbreak SmolLM2-\allowbreak 1.7B-\allowbreak Instruct} & 1.7 & \citep{allal2025smollm2smolgoesbig} & \texttt{rhymes-\allowbreak ai/\allowbreak Aria} & 25.3 & \citep{li2024aria} \\
\texttt{HuggingFaceTB/\allowbreak SmolLM2-\allowbreak 360M} & 0.4 & \citep{allal2025smollm2smolgoesbig} & \texttt{stabilityai/\allowbreak stablelm-\allowbreak 2-\allowbreak 1\_6b} & 1.6 & \citep{bellagente2024stablelm216b} \\
\texttt{HuggingFaceTB/\allowbreak SmolLM2-\allowbreak 360M-\allowbreak Instruct} & 0.4 & \citep{allal2025smollm2smolgoesbig} & \texttt{THUDM/\allowbreak glm-\allowbreak 4-\allowbreak 9b-\allowbreak hf} & 9.4 & \citep{glm2024chatglm} \\
\texttt{HuggingFaceTB/\allowbreak SmolVLM-\allowbreak Instruct} & 2.2 & \citep{marafioti2025smolvlm} & \texttt{tiiuae/\allowbreak falcon-\allowbreak 7b} & 7.2 & \citep{almazrouei2023falconseriesopenlanguage} \\
\texttt{ibm-\allowbreak granite/\allowbreak granite-\allowbreak 3.1-\allowbreak 2b-\allowbreak base} & 2.5 & \citep{granite2024granite3} & \texttt{tiiuae/\allowbreak Falcon3-\allowbreak 7B-\allowbreak Base} & 7.5 & \citep{tii2024falcon3} \\
\texttt{ibm-\allowbreak granite/\allowbreak granite-\allowbreak 3.1-\allowbreak 8b-\allowbreak base} & 8.2 & \citep{granite2024granite3} & \texttt{tiiuae/\allowbreak Falcon3-\allowbreak 7B-\allowbreak Instruct} & 7.5 & \citep{tii2024falcon3} \\
\texttt{ibm-\allowbreak granite/\allowbreak granite-\allowbreak 3.1-\allowbreak 8b-\allowbreak instruct} & 8.2 & \citep{granite2024granite3} & \texttt{TinyLlama/\allowbreak TinyLlama-\allowbreak 1.1B-\allowbreak intermediate-\allowbreak step-\allowbreak 1431k-\allowbreak 3T} & 1.1 & \citep{zhang2024tinyllama} \\
\texttt{llava-\allowbreak hf/\allowbreak llava-\allowbreak v1.6-\allowbreak mistral-\allowbreak 7b-\allowbreak hf} & 7.6 & \citep{liu2024improvedbaselinesvisualinstruction,liu2024llavanext} & \texttt{Zyphra/\allowbreak Zamba2-\allowbreak 1.2B} & 1.2 & \citep{glorioso2024zamba2suitetechnicalreport} \\
\texttt{LLM360/\allowbreak Amber} & 6.7 & \citep{liu2023llm360} &  &  &  \\
\bottomrule
\end{tabular}
\endgroup

\end{table}

\FloatBarrier

\section{\texorpdfstring{\tw{Per-token marginal attention}}{Per-token marginal attention}: additional evidence}
\label{app:agreement}

\subsection{Agreement by pair class}
\label{app:agreement:pairs}

The right part of \cref{fig:census_distribution} groups model pairs by what they share and gives the mean agreement of the \tw{per-token} profile in each class (850 natural documents, after removing position). A base model and its instruct version agree at $0.9684$ (12 pairs), and a vision-language model and the checkpoint it was built from at $0.9406$ (3 pairs). Two models of the same family and data at different sizes agree at $0.8894$ (22 pairs), and two pythia models of the same size that differ only in the pretraining seed at $0.8213$ (225 pairs), so post-training moves the \tw{per-token} profile less than a change of seed; the seed pairs are scored on a longer axis (median 5676 characters, compared with 4784.5), so their level is not directly comparable. The ten pythia-410m reruns agree with the average of the other nine at $0.9421$ (\cref{app:agreement:seeds}). The same architecture from different labs agrees at $0.7880$ (28 pairs) and different labs, data and tokenizers at $0.7551$ (85 pairs), a gap smaller than the spread within either class (standard deviation $0.060$ and $0.088$). Over all pairs of the census the mean agreement is $0.7607$ $[0.7556, 0.7658]$, below the $0.8673$ of one model compared with the consensus of the others, since averaging over models removes model-specific deviations.

\subsection{Agreement saturates at about twenty models}
\label{app:agreement:saturation}

Agreement with a consensus of $k$ models rises quickly with $k$ and then saturates. On the 850 natural documents, after removing position, the \tw{per-token marginal attention} of a model agrees with a consensus of $k$ other models at $0.7593$ for $k=1$, $0.8432$ for $k=5$, $0.8567$ for $k=10$, $0.8630$ for $k=20$, and $0.8673$ with the whole roster. The size of the roster is therefore not what produces the headline.

\subsection{A different document gives near-zero agreement}
\label{app:agreement:otherdoc}

Scored against the consensus computed on a different document, the \tw{per-token marginal attention} agrees at $0.0155$ with the other models and at $0.0139$ between two models (natural documents; $0.0128$ and $0.0114$ for the unweighted variant, $\omega \equiv 1$). The agreement of the main text is about the document at hand, not a generic shape shared by all attention profiles.

\subsection{Robustness: punctuation, whitespace and head weighting}
\label{app:agreement:robustness}

\begin{table}[t]
\centering
\caption{\textbf{Deleting punctuation and whitespace from the character axis does not lower agreement.} Each row deletes one class of characters from the axis, or keeps only that class, and compares the \tw{per-token marginal attention} of 67 models on 850 natural documents: one model with the consensus of the other 66, and two models with each other. Whitespace only is the one low row; it keeps a character only where every model's token is whitespace, which leaves 2.3\% of the axis and no character at all in 205 of the 850 documents. The two rows under the rule are references: a different document, and the unweighted variant ($\omega \equiv 1$) on all characters.}

\label{tab:census_deletion}
\begingroup
\footnotesize
\setlength{\tabcolsep}{4pt}
\begin{tabular}{l r r r r}
\toprule
characters on the axis & axis (\%) & documents & vs the other 66 & two models \\
\midrule
all characters & 100.0 & 850 & 0.8673 & 0.7607 \\
punctuation deleted & 90.8 & 850 & 0.8705 & 0.7664 \\
punctuation and whitespace deleted & 87.6 & 850 & 0.8734 & 0.7711 \\
punctuation only & 5.0 & 848 & 0.8648 & 0.7601 \\
whitespace only & 2.3 & 645 & 0.7336 & 0.6169 \\
\midrule
a different document (control) & 100.0 & 850 & 0.0155 & 0.0139 \\
unweighted ($\omega \equiv 1$), all characters & 100.0 & 850 & 0.8605 & 0.7491 \\
\bottomrule
\end{tabular}
\endgroup

\end{table}

\Cref{tab:census_deletion} removes classes of characters from the axis before ranking and scores the \tw{per-token marginal attention} on the 850 natural documents again. Deleting punctuation raises agreement from $0.8673$ to $0.8705$, and deleting punctuation and whitespace raises it to $0.8734$ with $87.6\%$ of the axis left. Punctuation alone, $5.0\%$ of the axis, gives $0.8648$. Whitespace alone is the only low row, at $0.7336$, and it covers $2.3\%$ of the axis. The unweighted variant ($\omega \equiv 1$) gives $0.8605$ on the same documents. On the 1000-document census with the unweighted variant, deleting punctuation and whitespace also raises agreement ($0.8459$, compared with $0.8396$ undeleted), but deleting punctuation alone lowers it slightly ($0.8361$) and keeping punctuation alone raises it ($0.8524$).

\subsection{The seed ceiling}
\label{app:agreement:seeds}

The seed ceiling is measured on 50 pythia pretraining-seed reruns, 10 at each of 5 sizes, never pooled across sizes. On the 850 natural documents, after removing position, each rerun agrees with the average of the other nine of its size at $0.7810$ (14M), $0.8674$ (31M), $0.9140$ (70M), $0.9491$ (160M) and $0.9421$ (410M). The census headline of $0.8673$ should be read against the two largest pools, not against $1$.

\subsection{Agreement on scrambled text}
\label{app:agreement:scrambled}

Within each tokenizer group, agreement orders the conditions as real text $>$ shuffled words $>$ shuffled characters $>$ random tokens (1000-document census, unweighted variant): in the pythia group $0.8277$ on real text, $0.7201$ on shuffled words and $0.5974$ on random token ids. The conditions are not length-matched, since the 2048-token cap spans fewer characters on scrambled text, so the gap sizes below the shuffled-words condition are provisional; the ordering of real text above shuffled words is not affected.

\FloatBarrier

\section{\texorpdfstring{\hw{Per-head marginal attention}}{Per-head marginal attention}: additional evidence}
\label{app:head_details}

\paragraph{Agreement by pair class.}
The right part of \cref{fig:head_write_distribution} groups model pairs by what they share, after averaging each \hw{per-head} profile over the 170 documents and binning it into ten relative-depth bins. Pairs in the class \textit{base vs derived model} (base and instruct, or a vision model and its backbone) agree at $0.9900$ (17 pairs), \textit{same family} (a different size or version) at $0.7829$ (55 pairs), \textit{same architecture, different family} at $0.8198$ (108 pairs), and \textit{different architecture} at $0.3728$ (2031 pairs), compared with $-0.0003$ for \textit{permuted heads}, where each model's heads are permuted before binning (2211 pairs; \cref{tab:head_write_main}). Same architecture agrees more than same family here, while for the \tw{per-token} profile the order is reversed (\cref{app:agreement:pairs}); the two sets of classes are built from different pair lists, so only their order is compared. Within one model (\textit{leave-one-document-out}), one document agrees with the average of the other 169 at $0.9529$ on the full vector over heads, without depth binning.

\paragraph{Two related statistics.} The per-head write in \cref{tab:head_write_main} is approximate: the stored data hold one value norm per layer, so every head of a layer is given the layer's mean value norm. Two further columns check this approximation from either side. The write per layer is exact but coarser, since it sums the heads of each layer; the output norm, the norm of each head's output, is exact per head but measures a different quantity. All three columns show a profile that is stable inside a model and much lower between architectures. The output norm is missing for one model, so its model and pair counts, given in parentheses, are smaller.

\begin{table}[htbp]
\centering
\caption{\textbf{Every head-wise statistic is stable inside one model and much lower between models of different architecture.} Rows are the comparisons of \cref{fig:head_write_distribution}; entries are the mean Spearman correlation between \hw{per-head} profiles over $n$ models or model pairs (in parentheses where the output norm has fewer), on 170 documents. Setting in \cref{app:head_details}.}

\label{tab:head_write_main}
\begingroup
\footnotesize
\setlength{\tabcolsep}{4pt}
\begin{tabular}{p{0.34\linewidth} r r r r}
\toprule
comparison & $n$ & \shortstack{write\\per head} & \shortstack{write\\per layer} & \shortstack{output\\norm} \\
\midrule
document split halves & 67 & 0.992 & 0.992 & 0.992 (66) \\
leave-one-document-out & 67 & 0.953 & 0.978 & 0.934 (66) \\
\midrule
base vs derived model & 17 & 0.990 & 0.990 & 0.987 \\
same family & 55 & 0.783 & 0.783 & 0.884 \\
same architecture, different family & 108 & 0.820 & 0.820 & 0.841 \\
different architecture & 2031 & 0.373 & 0.373 & 0.436 (1965) \\
permuted heads & 2211 & 0.000 & $-$0.005 & 0.001 (2145) \\
\bottomrule
\end{tabular}
\endgroup

\end{table}

\FloatBarrier

\section{Why marginal attention approximates the Jacobian}
\label{app:multilayer}

\subsection{A single attention layer}
\label{app:multilayer:single}

\begin{figure}[t]
\centering
\includegraphics[width=0.5\linewidth]{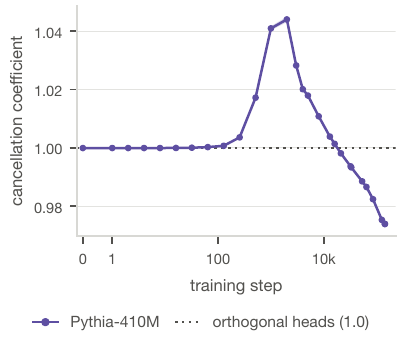}
\input{figures/main/head_cancellation_training_caption.tex}
\label{fig:head_cancellation_training}
\end{figure}

\begin{figure}[t]
\centering
\includegraphics[width=\linewidth]{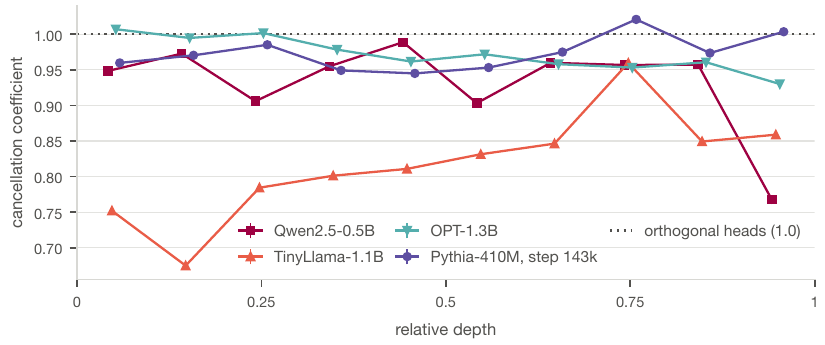}
\input{figures/main/head_cancellation_models_caption.tex}
\label{fig:head_cancellation_models}
\end{figure}

We now motivate this approximation heuristically. For a single causal attention layer, ignore the dependence of the attention weights on the input. Head $h$ adds $\sum_{s \le t} \attnel{h}{s}{t}\, W_{OV}^{(h)} \mathbf{x}_s$ to the residual stream at position $t$, with $W_{OV}^{(h)} := W_O^{(h)} W_V^{(h)}$, so its part of the Jacobian block is $[J_h^{(s)}]_t = \attnel{h}{s}{t} W_{OV}^{(h)}$, and the block is $[J^{(s)}]_t = \sum_h [J_h^{(s)}]_t$ up to the identity at $s = t$. Applied to the input it acts on, the head's part has norm
\begin{equation*}
    \bigl\lVert [J_h^{(s)}]_t\, \mathbf{x}_s \bigr\rVert = \attnel{h}{s}{t}\, \lVert W_O^{(h)} v_s \rVert = \wt{h}{s}{t}\, \attnel{h}{s}{t},
\end{equation*}
which is exactly the summand of \cref{defn:marginalAttention}. Summing over output positions and heads and normalising, \cref{defn:perToken,defn:marginalInfluence} become
\begin{equation*}
    \tokwel{s} \propto \sum_{t \ge s} \sum_{h} \bigl\lVert [J_h^{(s)}]_t\, \mathbf{x}_s \bigr\rVert,
    \qquad
    \inflel{s} \propto \sum_{t \ge s} \Bigl\lVert \sum_{h} [J_h^{(s)}]_t \Bigr\rVert_F,
\end{equation*}
with both normalisers summing the same expression over $s$. \tw{Per-token marginal attention} and \iw{per-token marginal influence} differ in two places. First, $\tokw$ sums the heads' norms where $\infl$ takes the norm of their sum, so they agree when the heads' contributions do not systematically cancel, for example when the heads write into linearly independent subspaces. \Cref{fig:head_cancellation_training,fig:head_cancellation_models,tab:head_cancellation} report this cancellation along the training of Pythia-410M and on four finished models. Second, $\tokw$ measures each head's part along the one direction $\mathbf{x}_s$ where $\infl$ takes its Frobenius norm over all directions; the two are proportional, with a factor that does not depend on $s$, when $\mathbf{x}_s$ is a generic direction for $W_{OV}^{(h)}$ and $\lVert \mathbf{x}_s \rVert$ is roughly constant, as it is after the normalisation that precedes attention. With $\omega \equiv 1$, the first-order term instead drops each head's gain $\lVert W_{OV}^{(h)} \mathbf{x}_s \rVert$ and keeps only the attention. This perhaps explains the lower correlation between the two quantities.

\begin{table}[t]
\centering

\caption{\textbf{Cancellation coefficient per model and block version.} We compute cancellation coefficients for the value channel alone, value channel plus the key channel, and with the query term on the diagonal: $\kappa$, head-summed block norm over the sum of per-head norms; its value for orthogonal heads; their ratio; the cross-term share (0 for orthogonal heads); the Spearman rank agreement over source positions between the sum-of-norms and norm-of-sum profiles; and the operator-norm ratio with its orthogonal-subspaces reference on 512 pairs per layer. Pythia-410M is shown at both initialization (step 0) and at the end of training.}

\label{tab:head_cancellation}
\begingroup
\footnotesize
\setlength{\tabcolsep}{4pt}
\begin{tabular}{llrrrrrrr}
\toprule
model & block & $\kappa$ & orth. & ratio & cross & rank & op. & op.\ orth. \\
\midrule
Qwen2.5-0.5B & value & 0.387 & 0.423 & 0.915 & $-$0.112 & 0.978 & 0.343 & 0.298 \\
 & value + key & 0.395 & 0.425 & 0.930 & $-$0.103 & 0.979 & 0.357 & 0.315 \\
 & value + key + query & 0.376 & 0.401 & 0.935 & $-$0.094 & 0.977 & -- & -- \\
\midrule
TinyLlama-1.1B & value & 0.216 & 0.258 & 0.834 & $-$0.386 & 0.960 & 0.166 & 0.141 \\
 & value + key & 0.226 & 0.265 & 0.852 & $-$0.373 & 0.959 & 0.218 & 0.189 \\
 & value + key + query & 0.221 & 0.258 & 0.858 & $-$0.360 & 0.954 & -- & -- \\
\midrule
OPT-1.3B & value & 0.283 & 0.295 & 0.959 & $-$0.189 & 0.975 & 0.213 & 0.186 \\
 & value + key & 0.458 & 0.459 & 0.997 & 0.001 & 0.977 & 0.487 & 0.414 \\
 & value + key + query & 0.448 & 0.449 & 0.997 & 0.001 & 0.976 & -- & -- \\
\midrule
Pythia-410M, step 0 & value & 0.262 & 0.262 & 1.000 & 0.000 & 0.990 & 0.138 & 0.106 \\
 & value + key & 0.262 & 0.262 & 1.000 & 0.000 & 0.990 & 0.117 & 0.111 \\
 & value + key + query & 0.262 & 0.262 & 1.000 & 0.000 & 0.990 & -- & -- \\
\midrule
Pythia-410M, step 143000 & value & 0.382 & 0.392 & 0.974 & $-$0.061 & 0.979 & 0.310 & 0.292 \\
 & value + key & 0.448 & 0.454 & 0.986 & $-$0.033 & 0.980 & 0.465 & 0.433 \\
 & value + key + query & 0.435 & 0.441 & 0.986 & $-$0.030 & 0.979 & -- & -- \\
\bottomrule
\end{tabular}
\endgroup

\end{table}

\subsection{Several layers}
\label{app:multilayer:multi}

\Cref{app:multilayer:single} argues for a single attention layer. To understand why this might continue to hold across multiple layers, write the input-output Jacobian as
\begin{equation*}
J = \prod_{\ell=1}^L (I + J_\ell) = I + \sum_{\ell=1}^L J_\ell + \text{(higher-order terms)}
\end{equation*}
where $J_\ell$ denotes the nontrivial Jacobian contribution of layer $\ell$. This expansion can be interpreted as a sum over paths through token space: the first-order terms are direct paths from $s$ to $t$ consisting of a single attention edge, and thus inherit the structure of the attention matrices, while terms such as $J_kJ_\ell$ correspond to two-hop paths $s \to u \to t$, weighted by attention coefficients, with successively higher-order terms corresponding to progressively longer paths. Under the assumption that these multi-hop paths distribute attention somewhat evenly, this leads to
\begin{equation*}
    J = I + \sum_{\ell=1}^L \sum_h A^{(\ell,h)} \otimes W_{OV}^{(\ell,h)}
    + \text{(multi-hop terms)}
\end{equation*}
with the first-order argument of \cref{sec:theory:approx} applying to each layer's heads, and the \tw{per-token marginal attention} summing their terms over all layers.

Concretely, for the two-layer case $s \rightarrow u \rightarrow t$, consider the distribution over $(s, u)$ induced by the first layer's attention pattern, $\mathbf{P}(s, u) \propto A_{s,u}^{(1)}$. Then, ``attention distributed evenly'' means $\mathbb{E}_{u}[a_u^{(2)} \mid s] = 1$ a.s., where $a_u^{(2)} := \sum_t A^{(2)}_{u,t}$ is the total attention paid to $u$ in the second layer, $n$ times its marginal attention with $\omega \equiv 1$. The condition is impossible under causal masking. However, violations are plausibly monotonic in position due to causal masking, perhaps explaining why higher-order terms disagree little with one another.

Heuristically, the multi-hop gain is also small. In general, only $\lVert W_{OV}^{(j)} W_{OV}^{(k)} \rVert_F \leq \lVert W_{OV}^{(j)} \rVert_F \lVert W_{OV}^{(k)} \rVert_{\mathrm{op}}$ holds. Suppose instead that the column space of $W_{OV}^{(k)}$ is randomly oriented relative to the row space of $W_{OV}^{(j)}$, that is, $W_{OV}^{(k)}$ is replaced by $Q W_{OV}^{(k)}$ with $Q$ a uniformly random rotation of $\mathbb{R}^{d_\text{model}}$. Then $\mathbb{E}_Q \lVert W_{OV}^{(j)} Q W_{OV}^{(k)} \rVert_F^2 = \lVert W_{OV}^{(j)} \rVert_F^2 \lVert W_{OV}^{(k)} \rVert_F^2 / d_\text{model}$, and since $W_{OV}^{(k)}$ has rank at most $d_\text{head}$, this is at most $(d_\text{head}/d_\text{model}) \lVert W_{OV}^{(j)} \rVert_F^2 \lVert W_{OV}^{(k)} \rVert_{\mathrm{op}}^2$. Under this approximation, the two-hop gain is smaller than the general bound by a factor of about $\sqrt{d_\text{head}/d_\text{model}}$, which plausibly further reduces the weight of the multi-hop terms in the Jacobian norm. Trained heads that compose are not randomly oriented, so this is not a bound.

\section{Theory: measurements}
\label{app:theory_measurements}

\Cref{tab:jacobian_census_headline} compares, over 16 models, how well the \iw{per-token marginal influence} and the \tw{per-token marginal attention} of each model agree with the consensus of the other models.

\begin{table}[t]
\centering
\caption{\textbf{Marginal influence is shared across models, marginal attention more so.} 16 models from 6 families, 100 documents. Spearman correlation per document of one model's map with the consensus of the other models, with its own family left out of the consensus, and between two models, each beside a different-document reference. Rows: the unweighted \tw{per-token marginal attention} and three row-gradient estimates of the \iw{per-token marginal influence} (\cref{app:theory_measurements:estimators}): gradient length, substitution sensitivity, gradient times input.}

\label{tab:jacobian_census_headline}
\begingroup
\footnotesize
\setlength{\tabcolsep}{3pt}
\begin{tabular}{l r r r r r}
\toprule
 & \multicolumn{2}{c}{vs all other models} & \multicolumn{2}{c}{vs other families} & \\
\cmidrule(lr){2-3}\cmidrule(lr){4-5}
map & agreement & different document & agreement & different document & two models \\
\midrule
unweighted profile & 0.8612 & 0.0334 & 0.8076 & 0.0321 & 0.7652 \\
influence, gradient length & 0.7722 & 0.0336 & 0.6768 & 0.0316 & 0.6295 \\
influence, substitution & 0.7759 & 0.0328 & 0.6828 & 0.0308 & 0.6349 \\
influence, gradient $\times$ input & 0.6568 & 0.0344 & 0.5797 & 0.0318 & 0.4771 \\
\bottomrule
\end{tabular}
\endgroup

\end{table}

\subsection{Estimators of the marginal influence}
\label{app:theory_measurements:estimators}

Two estimators of the \iw{per-token marginal influence} appear in \cref{sec:theory}. The \textbf{block norm} is the object of \cref{defn:marginalInfluence}: the Frobenius norm of each block of the Jacobian of the output logits with respect to the input embeddings, summed over output positions. We compute it only for Qwen2.5-0.5B, with Hutchinson's estimator on a sample of output positions in each document. Its agreement with the \tw{per-token marginal attention} is moderate (0.64), and lower with uniform weights $\omega \equiv 1$ (0.53).

The \textbf{row gradient} replaces the block norm where it is too costly to compute for every model. The influence of input position $s$ on output position $t$ is the norm of the gradient of the log-probability of the observed next token with respect to the embedding at $s$. For the observed token $y$ and the predicted distribution $\mathbf{p}_t$, this gradient is $(\mathbf{e}_y - \mathbf{p}_t)^\top [J^{(s)}]_t$, a probability-weighted combination of the rows of the block $[J^{(s)}]_t$, not one of its rows. Each output position's profile is normalised to sum to one over input positions, and at each input position $s$ the profiles are averaged over the sampled output positions $t \ge s$, so the number of profiles averaged depends on $s$. It is a proxy for the block norm, not the quantity of \cref{defn:marginalInfluence}; \cref{tab:jacobian_census_headline} also reports two variants of it. Most of the 16 models belong to two families, pythia and Qwen2.5, so this comparison covers fewer families than the census.

\subsection{Empirical diagnostics of the assumptions}
\label{app:theory_measurements:constants}

We report empirical diagnostics of the three assumptions and the logit gap on the tokenizer-sharing pairs and the transformers of \cref{sec:theory:identifiability}, with the same documents. Smoothness and current-token dependence are measured on sampled Jacobian blocks; the lag profile, which is too costly as a block for every model, uses the row gradient. These measurements do not establish the uniform bounds required by the theorem, and we do not use them to evaluate its quantitative bound.

\textbf{Logit gap.} The logit gap $\delta$ of a pair is the norm of the difference between the two models' centred next-token logits. As it grows, the agreement of the two models' \iw{per-token marginal influence} falls, and it orders the 26 pairs tightly (Spearman $-0.88$). Pooled over tokenizer groups it does not order the agreement of the \tw{per-token marginal attention}, which falls with the gap only inside each group.

\textbf{Smoothness.} We measure local sensitivity at each model's own input embeddings. We move one embedding by a small step in a random direction and divide the relative change of the sampled Jacobian block by the relative size of the step. This ratio is of order one for every model and changes little with the step for most of them, consistent with approximately linear changes over the sampled directions and step sizes. It is lower for larger models (Spearman $-0.62$ with parameter count), indicating lower relative local sensitivity on this diagnostic. This dimensionless ratio is distinct from the uniform, absolute smoothness constant $M$ in \cref{assump:jacobianSmoothness}.

\textbf{Current-token dependence.} The constant $c$ of \cref{assump:preTokenDependence} bounds from below the Jacobian block that links each prediction to the last token it reads, the token at its own position. The block is above zero at every sampled position of every model, and it holds a sizeable share of the block norm of its row (median 0.22 to 0.48 across models). The same holds for the row gradient of the models without attention (\cref{app:theory_measurements:attention_free}). The constant is in logit units and differs across models by orders of magnitude, but the theorem only needs each model's own $c$ to be above zero.

\textbf{Lag decay.} \Cref{assump:lagDecay} asks that the dependence of a prediction on earlier tokens fall toward zero with their distance. This dependence decays as a power law of the distance, with an exponent of about one half to two thirds, and the models without attention decay in the same way. Larger models decay more slowly. The decay is slow: between a fifth and a third of the dependence lies more than 512 tokens back.

\subsection{Emergence during training}
\label{app:theory_measurements:emergence}

\begin{figure}[t]
\centering
\includegraphics[width=0.5\linewidth]{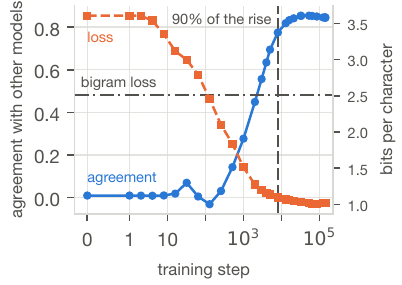}
\input{figures/main/emergence_caption.tex}
\label{fig:emergence}
\end{figure}

\textbf{Experimental setting.} We follow one model through training to ask whether its agreement with other models grows as its language-modelling loss falls. The subject is Pythia-410M at 28 public training checkpoints, from random initialisation (step 0) to the final model (step 143{,}000), on 60 Pile passages cut at 512 tokens (median scored length 2192 characters). At each checkpoint we compute the unweighted ($\omega \equiv 1$) \tw{per-token marginal attention} and score it with the estimator of \cref{sec:census}: projection onto the characters, ranks, residualisation on 50 relative-position bins, and Spearman correlation per document. The reference is the mean of the same profile over four fully trained models of other families (Qwen2.5-0.5B, Qwen2.5-1.5B, SmolLM2-360M and OLMo-2-1B), so Pythia is never part of its own target. Agreement is this correlation averaged over documents, and the loss is the checkpoint's bits per character on the same passages, from the same forward pass. As a loss reference, we fit a bigram model with absolute discounting, backed off to a unigram model, on the other 440 passages of the corpus, and score it on the same 60 passages with the same tokenizer. We ran no bootstrap, so we report no interval.

\textbf{Result analysis.} Agreement is near zero at random initialisation and high at the final step (0.84), so the model acquires the shared profile during training and does not have it at the start. Most of the rise is early: it is nearly complete by step 8{,}000, while the loss is still falling. Agreement stays near zero as long as the loss is above the bigram model's 2.51 bits per character, and rises only after the loss drops below it, so the model shares the profile only once it predicts better than token-pair statistics. The evidence is narrow: one model, one training seed, and 60 documents of at most 512 tokens. The loss also stands in for the logit gap $\delta$ of \cref{thm:jacobianControl}, which we do not compute between this model and the four reference models.

\subsection{Models without attention}
\label{app:theory_measurements:attention_free}

\begin{figure}[t]
\centering
\includegraphics[width=0.5\linewidth]{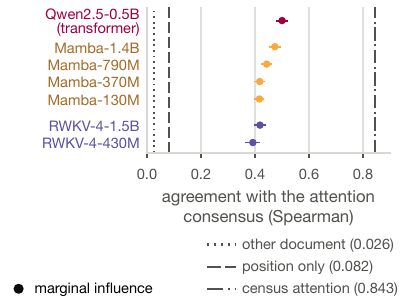}
\input{figures/main/attention_free_def4_caption.tex}
\label{fig:attention_free_def4}
\end{figure}

\textbf{Experimental setting.} We ask whether the shared \tw{per-token marginal attention} comes from the attention mechanism or from next-token prediction, by computing the \iw{per-token marginal influence} of models that have no attention. The subjects are four Mamba models \citep[130M to 1.4B parameters;][]{gu2024mamba} and two RWKV-4 models \citep[430M and 1.5B;][]{peng-etal-2023-rwkv}, with Qwen2.5-0.5B as the one transformer reference. The influence is the object of \cref{defn:marginalInfluence}, estimated with Hutchinson's method (8 probes) on 128 sampled output positions per document, all at position 64 or later, at 2048 tokens; RWKV-4 reads at most 1024 tokens, so its rows use the first 1024. The reference is the consensus of the unweighted ($\omega \equiv 1$) \tw{per-token marginal attention} of the census models on the same document, and each influence map is compared with it using the estimator of \cref{sec:census}; Qwen2.5-0.5B's own attention is part of that consensus. We use 100 documents of the census corpus, five per domain, the three generated domains included, and \cref{fig:attention_free_def4} shows bootstrap intervals over documents.

\textbf{Result analysis.} The influence of the four Mamba models agrees with the attention consensus nearly as well as the transformer's does (0.42 to 0.47, and 0.50 for Qwen2.5-0.5B), so the shared profile does not require the attention mechanism. Every model stays far above a map of position alone and above its own influence map compared with a different document. The evidence is narrow: one transformer reference, documents that include generated text, and no logit gap $\delta$ for any model without attention, so this is not a test of \cref{thm:jacobianControl}.

\subsection{What the shared per-token profile encodes}
\label{app:theory_measurements:predictors}

\textbf{Experimental setting.} To ask what the shared \tw{per-token marginal attention} encodes, we predict it from properties of the text alone. The target is the consensus per-token profile of the census, unweighted ($\omega \equiv 1$), per word, ranked and residualised on 50 relative-position bins as in \cref{sec:census}. The predictors are five groups of word properties: position in the document; surface form (length, punctuation, capitalisation, sentence edges); word frequency (external and in-corpus counts); local predictability (surprisal and entropy at the word and the next token); and grammatical role (parse depth, number of dependents, part of speech, dependency label). We fit a ridge regression on each group alone and on all five together, out of fold by document (5 folds), and score it by Spearman correlation per document over 550 documents of 11 English-prose domains. The reference is one model's own per-token profile compared with the consensus of the other 66, averaged over the 67 census models.

\textbf{Result analysis.} The five groups together predict the consensus profile only in part (0.52), well below a model's own unweighted per-token profile (0.88). Alone, grammatical role predicts it best, then surface form, local predictability and word frequency, and position predicts nothing (right panel of \cref{fig:jacobian_predictors}). Removing one group from the five costs the most for grammatical role and local predictability, little for surface form and word frequency, and nothing for position.

\FloatBarrier

\section{KV eviction details}
\label{app:kv_details}

\subsection{Head budget details}
\label[appendix]{app:kv_details:budget}

The set $\mathcal{D}_{\text{pre}}$ of \cref{eq:budget} holds 200 documents from the Nemotron pretraining sample \citep{nemotron}, the calibration corpus of KVzap \citep{jegou2026kvzapfastadaptivefaithful}, and is disjoint from every benchmark.
KVzip+ \citep{jegou2026kvzapfastadaptivefaithful} scores every entry $(h,s)$ of a document by the largest value-weighted attention it receives from the queries $\mathcal{R}(X)$ of KVzip's reconstruction pass, normalised by the attention input $h_t$, and keeps the fraction $B$ of all entries above one global cut $\tau_B(X)$. The fraction $m_h(X)$ behind the share $p_h(X)$ of \cref{eq:budget} counts head $h$'s own entries above that cut:
\begin{equation}
    \label{eq:headmass_share}
    \begin{aligned}
    m^{\mathrm{rec}}_{h,s}(X) &= \max_{t \in \mathcal{R}(X)} \frac{\attnel{h}{s}{t}(X)\,\bigl\lVert W_O^{(h)} v_s(X) \bigr\rVert}{\lVert h_t \rVert},
    \qquad
    \frac{1}{Hn}\sum_{j=1}^{H}\sum_{s=1}^{n} \mathbf{1}\bigl[m^{\mathrm{rec}}_{j,s}(X) > \tau_B(X)\bigr] = B,\\
    m_h(X) &= \frac{1}{n}\sum_{s=1}^{n} \mathbf{1}\bigl[m^{\mathrm{rec}}_{h,s}(X) > \tau_B(X)\bigr].
    \end{aligned}
\end{equation}

\begin{table}[htbp]
\centering
\caption{\textbf{\headmass{} tables from disjoint sets of pretraining documents agree at a median Spearman of 0.98 or higher.} Spearman correlation over each target's key-value heads between the per-head marginal attention (\cref{eq:budget}) of two disjoint sets of the 200 pretraining documents, read on KVzip's reconstruction queries at 8$\times$. \emph{1 vs 199} is the median (5th percentile) over documents, \emph{5 vs 195} and \emph{20 vs 180} the median of 100 draws, and \emph{100 vs 100} the even documents compared with the odd ones. The \emph{top decile} column is the Jaccard overlap of the top tenth of heads between even and odd documents, and the \emph{prefill queries} column is the 1 vs 199 median on the document's own queries.}

\label{tab:kv_budget_stability}
\begingroup
\footnotesize
\setlength{\tabcolsep}{4pt}
\begin{tabular}{lrrrrrrr}
\toprule
target & \shortstack[r]{key-value\\heads} & 1 vs 199 & 5 vs 195 & 20 vs 180 & 100 vs 100 & top decile & prefill queries \\
\midrule
Qwen3-8B & 288 & 0.9828~{\scriptsize (0.9704)} & 0.9952 & 0.9985 & 0.9995 & 1.0000 & 0.9624 \\
Llama-3.1-8B & 256 & 0.9830~{\scriptsize (0.9744)} & 0.9951 & 0.9985 & 0.9994 & 0.9259 & 0.9394 \\
phi-4 & 400 & 0.9836~{\scriptsize (0.9706)} & 0.9954 & 0.9986 & 0.9992 & 0.9512 & 0.9681 \\
Qwen3-32B & 512 & 0.9815~{\scriptsize (0.9512)} & 0.9952 & 0.9986 & 0.9993 & 1.0000 & 0.9656 \\
Mistral-7B & 256 & 0.9829~{\scriptsize (0.9729)} & 0.9954 & 0.9984 & 0.9994 & 1.0000 & 0.9666 \\
\bottomrule
\end{tabular}
\endgroup

\end{table}

\begin{figure}[t]
\centering
\includegraphics[width=\linewidth]{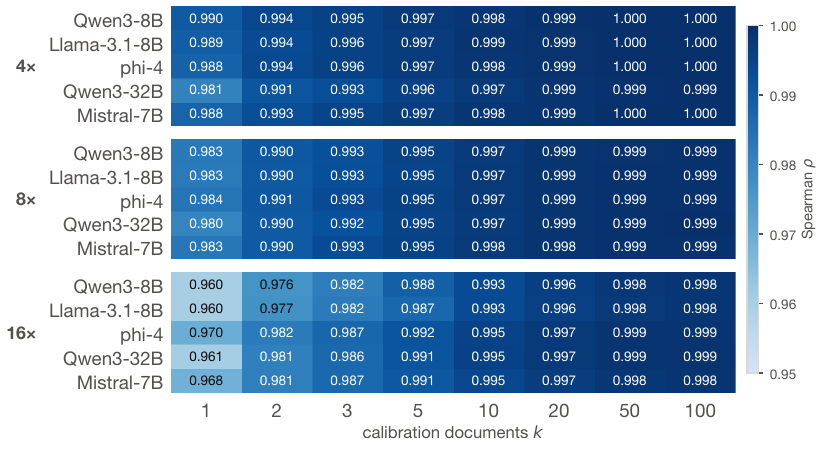}
\input{figures/main/kv_budget_documents_caption.tex}
\label{fig:kv_budget_documents}
\end{figure}

\begin{table}[htbp]
\centering
\caption{\textbf{LU-KV, KVzip+ and Proportional agree with \headmass{}, prefill attention doesn't.} Spearman correlation with the \headmass{} table over each target's (layer, key-value head) grid at 8$\times$, for the tables of \cref{app:kv_details:budget_maps}. The last row is the mean over targets.}

\label{tab:kv_budget_correlations}
\begingroup
\footnotesize
\setlength{\tabcolsep}{3pt}
\begin{tabular}{lrrrrr}
\toprule
target & LU-KV & KVzip+ & Proportional & Prefill attention & key-value heads \\
\midrule
Qwen3-8B & 0.81 & 0.92 & 0.96 & 0.13 & 288 \\
Llama-3.1-8B & 0.80 & 0.91 & 0.95 & 0.38 & 256 \\
phi-4 & 0.81 & 0.92 & 0.97 & 0.30 & 400 \\
Qwen3-32B & 0.80 & 0.91 & 0.98 & 0.17 & 512 \\
Mistral-7B & 0.81 & 0.93 & 0.96 & 0.42 & 256 \\
\midrule
mean over targets & 0.81 & 0.92 & 0.96 & 0.28 & -- \\
\bottomrule
\end{tabular}
\endgroup

\end{table}

\subsection{Head budget maps, one per target}
\label[appendix]{app:kv_details:budget_maps}

One figure per target, at $8\times$: the \headmass{} table, LU-KV's calibrated budget, KVzip+'s per-head shares, the proportional budget (\cref{eq:budget_write}) and ordinary prefill attention, head by head. The \headmass{} table agrees closely, head for head, with the keep rates KVzip+ realises on held-out documents (Spearman 0.91 to 0.93, \cref{tab:kv_budget_correlations}), so averaging over pretraining text loses little. The same recipe applied to the reading queries, the document's own positions, agrees with it far less, as the maps show.

\begin{figure}[t]
\centering
\includegraphics[width=\linewidth]{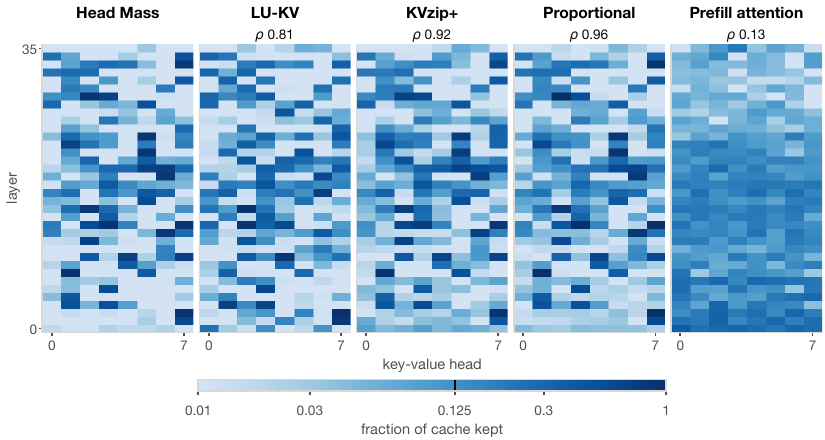}
\input{figures/main/kv_head_budget_maps_qwen3-8b_caption.tex}
\label{fig:kv_head_budget_maps_qwen3-8b}
\end{figure}

\begin{figure}[t]
\centering
\includegraphics[width=\linewidth]{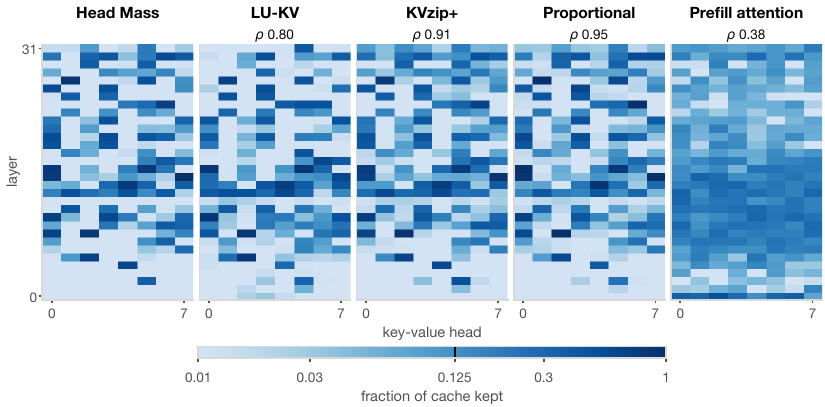}
\input{figures/main/kv_head_budget_maps_llama-31-8b_caption.tex}
\label{fig:kv_head_budget_maps_llama-31-8b}
\end{figure}

\begin{figure}[t]
\centering
\includegraphics[width=\linewidth]{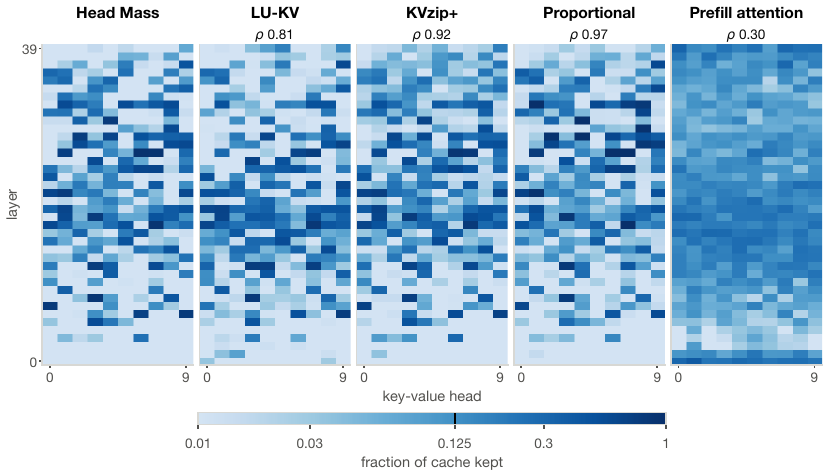}
\input{figures/main/kv_head_budget_maps_phi-4_caption.tex}
\label{fig:kv_head_budget_maps_phi-4}
\end{figure}

\begin{figure}[t]
\centering
\includegraphics[width=\linewidth]{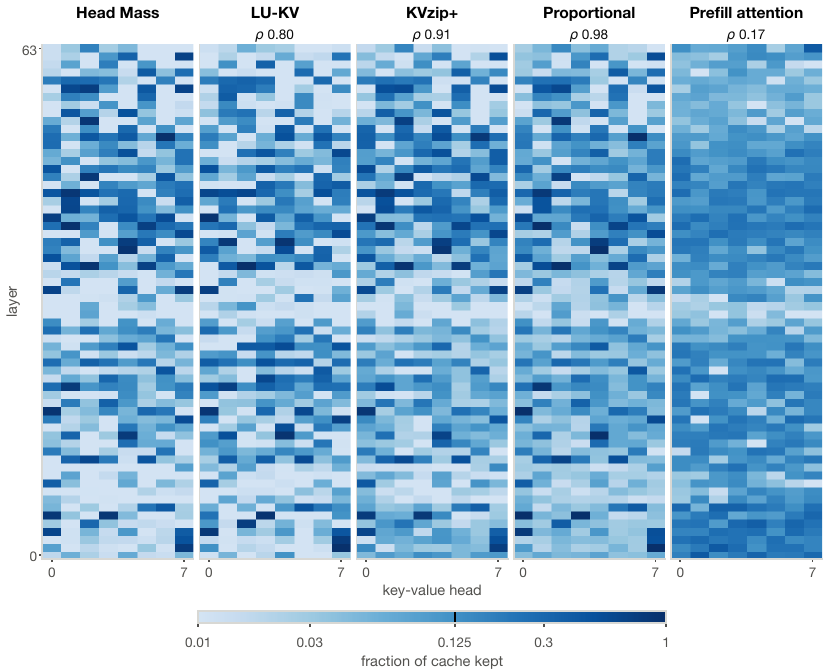}
\input{figures/main/kv_head_budget_maps_qwen3-32b_caption.tex}
\label{fig:kv_head_budget_maps_qwen3-32b}
\end{figure}

\begin{figure}[t]
\centering
\includegraphics[width=\linewidth]{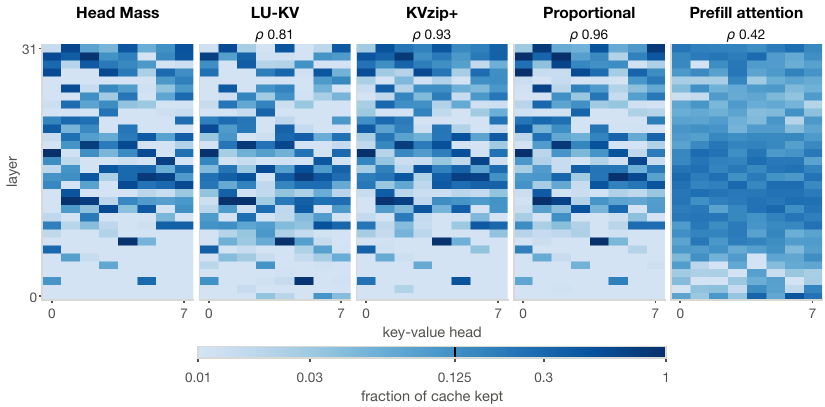}
\input{figures/main/kv_head_budget_maps_mistral-7b_caption.tex}
\label{fig:kv_head_budget_maps_mistral-7b}
\end{figure}

\subsection{Proportional budget ablation}
\label[appendix]{app:kv_details:write}

The ablation replaces \headmass{}'s threshold share with a share proportional to each head's summed KVzip+ score on the reconstruction pass,
\begin{equation}
    \label{eq:budget_write}
    \hw{\beta_h} \;\propto\; \mathbb{E}_{X\sim\mathcal{D}_{\text{pre}}}\Bigl[\,\sum_{s} \max_t \attnel{h}{s}{t}(X)\,\lVert W_O^{(h)} v_s\rVert / \lVert h_t\rVert\Bigr],
\end{equation}
max-pooled over each key-value group, then clipped and rescaled as in \cref{eq:budget}. With EA as the scorer it stays within half a point of \hw{HM}~+~EA on Qwen3-8B at every setting, trails it by about one to three points on the other four targets at most compression ratios, and trails it by more than ten points on Qwen3-32B at $16\times$ (\cref{tab:kv_write_ablation}). On Qwen3-8B the cut costs under half a point, while reading the same budget on the reading queries instead of the reconstruction pass costs over 30 points at every setting (\cref{tab:kv_write_equivalence}).

\begin{table}[htbp]
\centering
\caption{\textbf{HM + EA needs one offline budget table per target and no work per document.} Work each method needs beyond ordinary serving. $\triangle$ marks work done once per target and $\bullet$ work per document; two marks denote training a surrogate or an extra target forward. HM is \headmass{} and EA is Expected Attention. The last column is the wall time of the offline step on Mistral-7B-Instruct-v0.3 (cluster job walltime).}

\label{tab:kv_cost}
\begingroup
\footnotesize
\setlength{\tabcolsep}{1.6pt}
\begin{tabular}{>{\raggedright\arraybackslash}m{1.0in}c>{\columncolor{kvoffline!10}\raggedright\arraybackslash}p{1.4in}>{\columncolor{kvonline!10}\raggedright\arraybackslash}p{1.4in}>{\raggedright\arraybackslash}p{0.95in}}
\toprule
 & & \textbf{offline, per target} & \textbf{online, per document} & \textbf{offline time, Mistral-7B} \\
\midrule
AdaKV + EA &  & none & none & none \\
AdaKV + KVzap & \textcolor{kvoffline}{$\triangle\triangle$} & surrogate training on KVzip's score & surrogate call, not counted & 4 h 14 m, one H100 \\
KVzip & \textcolor{kvonline}{$\bullet\bullet$} & none & one extra target forward & none \\
LU-KV + EA & \textcolor{kvoffline}{$\triangle$} & per-head budget curve calibration & none & published curve \\
\hw{HM} + EA & \textcolor{kvoffline}{$\triangle$} & per-head budget table, 200 pretraining documents & none & 34 min \\
\hw{HM} + Guide & \textcolor{kvoffline}{$\triangle$}\,\textcolor{kvonline}{$\bullet$} & the same table and a guide map & one Qwen2.5-0.5B-Instruct forward & 36 min \\
\bottomrule
\end{tabular}
\endgroup

\end{table}

\begin{table}[t]
\centering
\caption{\textbf{With EA as the scorer, the proportional per-head budget (\cref{eq:budget_write}) is within half a point of HM + EA at every setting only on Qwen3-8B; elsewhere it trails by up to 3.3 points, except on Qwen3-32B at 16$\times$, where it trails by more than 11.} Each cell is the RULER score of the proportional budget with EA; in parentheses, its paired difference from HM + EA, bold where the 95\% interval excludes zero. Both budgets come from KVzip's reconstruction pass on the same pretraining documents.}

\label{tab:kv_write_ablation}
\begingroup
\footnotesize
\setlength{\tabcolsep}{3pt}
\begin{tabular}{lrrrrr}
\toprule
 & \multicolumn{3}{c}{RULER, context 4096} & \multicolumn{2}{c}{RULER, context 8192} \\
\cmidrule(lr){2-4}\cmidrule(lr){5-6}
Target & 4$\times$ & 8$\times$ & 16$\times$ & 8$\times$ & 16$\times$ \\
\midrule
Qwen3-8B & 94.8 (\textbf{$-$0.2}) & 91.9 (0.2) & 73.1 ($-$0.2) & 89.1 (0.1) & 69.5 (0.4) \\
Llama-3.1-8B & 95.0 ($-$0.1) & 90.8 (\textbf{$-$1.5}) & 65.1 (\textbf{$-$1.2}) & 86.3 (\textbf{$-$1.7}) & 60.7 (\textbf{$-$2.1}) \\
phi-4 & 90.8 (\textbf{$-$0.6}) & 84.8 (\textbf{$-$3.0}) & 66.8 (\textbf{$-$1.7}) & 82.8 (\textbf{$-$2.1}) & 67.3 (\textbf{$-$1.1}) \\
Qwen3-32B & 95.0 (\textbf{$-$0.2}) & 92.0 (\textbf{$-$1.2}) & 67.6 (\textbf{$-$11.5}) & 90.9 (\textbf{$-$1.1}) & 66.5 (\textbf{$-$11.4}) \\
Mistral-7B & 89.3 (\textbf{$-$1.2}) & 83.2 (\textbf{$-$1.1}) & 63.2 (\textbf{$-$2.2}) & 79.7 (\textbf{$-$0.7}) & 57.8 (\textbf{$-$3.3}) \\
\bottomrule
\end{tabular}
\endgroup

\end{table}

\begin{table}[t]
\centering
\caption{\textbf{On Qwen3-8B, the proportional per-head budget (\cref{eq:budget_write}) read on the reconstruction queries is within 0.25 points of HM + EA; read on the document's own queries it scores below AdaKV + EA at every compression ratio.} Over heads, the Spearman correlation between the reading-pass budget and \headmass{} is $-$0.11, $-$0.15 and $-$0.10 at 4$\times$, 8$\times$ and 16$\times$. The upper rows are RULER-4096 scores (mean over 13 tasks, 6500 documents); the lower rows are paired differences with 95\% bootstrap intervals.}

\label{tab:kv_write_equivalence}
\begingroup
\footnotesize
\setlength{\tabcolsep}{3pt}
\begin{tabular}{l r r r}
\toprule
 & 4$\times$ & 8$\times$ & 16$\times$ \\
\midrule
AdaKV + EA & 88.4 & 72.8 & 46.5 \\
\hw{HM} + EA & 95.0 & 91.8 & 73.3 \\
\hw{HM} + KVzap & 94.9 & 92.0 & 73.0 \\
Proportional (reconstruction) + EA & 94.8 & 91.9 & 73.1 \\
Proportional (reading) + EA & 61.1 & 53.0 & 37.7 \\
\midrule
\multicolumn{4}{l}{\emph{paired difference, with its 95\% interval}} \\
Proportional (reconstruction) $-$ HM + EA & $-$0.21 & 0.15 & $-$0.25 \\
 & [$-$0.40, $-$0.04] & [$-$0.22, 0.50] & [$-$1.01, 0.47] \\
\addlinespace
Proportional (reading) $-$ HM + EA & $-$33.94 & $-$38.72 & $-$35.59 \\
 & [$-$34.52, $-$33.31] & [$-$39.43, $-$38.00] & [$-$36.47, $-$34.68] \\
\addlinespace
Proportional (reading) $-$ AdaKV + EA & $-$27.37 & $-$19.78 & $-$8.73 \\
 & [$-$28.07, $-$26.62] & [$-$20.58, $-$19.06] & [$-$9.64, $-$7.82] \\
\bottomrule
\end{tabular}
\endgroup

\end{table}

\begin{table}[htbp]
\centering
\caption{\textbf{At context 8192, HM + EA scores above AdaKV + EA in all 10 columns.} Mean string-match score over the 13 RULER tasks at context 8192, 6500 documents and one run per cell, for five targets at 8$\times$ and 16$\times$ compression. Paired 95\% bootstrap half-widths of the differences from HM + Guide are 0.33 to 1.09 points. Methods, shading, cost symbols and italics as in \cref{tab:kv_ruler_4096}; the uncompressed row is not ranked.}

\label{tab:kv_ruler_8192}
\begingroup
\footnotesize
\setlength{\tabcolsep}{1.6pt}
\ifdefined\kvshadecolw\else\newlength{\kvshadecolw}\fi
\settowidth{\kvshadecolw}{\footnotesize\bfseries Llama-3.1-8B}
\addtolength{\kvshadecolw}{-2\tabcolsep}
\divide\kvshadecolw by 2
\ifdefined\kvshadecolwfl\else\newlength{\kvshadecolwfl}\fi
\settowidth{\kvshadecolwfl}{\footnotesize\bfseries 71.7}
\ifdim\kvshadecolw<\kvshadecolwfl\setlength{\kvshadecolw}{\kvshadecolwfl}\fi
\begin{tabular}{>{\raggedright\arraybackslash}m{0.9in}cc>{\centering\arraybackslash}m{\kvshadecolw}>{\centering\arraybackslash}m{\kvshadecolw}>{\centering\arraybackslash}m{\kvshadecolw}>{\centering\arraybackslash}m{\kvshadecolw}>{\centering\arraybackslash}m{\kvshadecolw}>{\centering\arraybackslash}m{\kvshadecolw}>{\centering\arraybackslash}m{\kvshadecolw}>{\centering\arraybackslash}m{\kvshadecolw}>{\centering\arraybackslash}m{\kvshadecolw}>{\centering\arraybackslash}m{\kvshadecolw}}
\toprule
 & \multicolumn{2}{c}{\textbf{cost}} & \multicolumn{2}{c}{\textbf{Qwen3-8B}} & \multicolumn{2}{c}{\textbf{Llama-3.1-8B}} & \multicolumn{2}{c}{\textbf{phi-4}} & \multicolumn{2}{c}{\textbf{Qwen3-32B}} & \multicolumn{2}{c}{\textbf{Mistral-7B}} \\
\cmidrule(lr){2-3}\cmidrule(lr){4-5}\cmidrule(lr){6-7}\cmidrule(lr){8-9}\cmidrule(lr){10-11}\cmidrule(lr){12-13}
 & \scriptsize offline & \scriptsize online & 8$\times$ & 16$\times$ & 8$\times$ & 16$\times$ & 8$\times$ & 16$\times$ & 8$\times$ & 16$\times$ & 8$\times$ & 16$\times$ \\
\midrule
AdaKV + EA &  &  & \cellcolor{kvshade!12}71.7 & \cellcolor{kvshade!30}50.9 & \cellcolor{kvshade!12}41.2 & \cellcolor{kvshade!12}15.8 & \cellcolor{kvshade!12}41.7 & \cellcolor{kvshade!12}13.6 & \cellcolor{kvshade!12}64.1 & \cellcolor{kvshade!12}38.4 & \cellcolor{kvshade!12}33.5 & \cellcolor{kvshade!12}20.5 \\
AdaKV + KVzap & $\triangle\triangle$ &  & \cellcolor{kvshade!47}86.6 & \cellcolor{kvshade!12}46.5 & \cellcolor{kvshade!30}86.7 & \cellcolor{kvshade!30}60.5 & \cellcolor{kvshade!30}\textit{70.8} & \cellcolor{kvshade!30}\textit{43.9} & \cellcolor{kvshade!47}88.5 & \cellcolor{kvshade!47}50.3 & \cellcolor{kvshade!30}\textit{67.0} & \cellcolor{kvshade!30}\textit{40.2} \\
KVzip &  & $\bullet\bullet$ & \cellcolor{kvshade!100}\textbf{89.8} & \cellcolor{kvshade!47}55.6 & \cellcolor{kvshade!100}\textbf{90.1} & \cellcolor{kvshade!100}\textbf{73.6} & \cellcolor{kvshade!100}\textbf{88.0} & \cellcolor{kvshade!47}64.2 & \cellcolor{kvshade!100}\textbf{95.0} & \cellcolor{kvshade!100}\textbf{92.1} & \cellcolor{kvshade!100}\textbf{82.5} & \cellcolor{kvshade!100}\textbf{66.3} \\
LU-KV + EA & $\triangle$ &  & \cellcolor{kvshade!30}\textit{72.9} & \cellcolor{kvshade!65}\textit{55.7} & \cellcolor{kvshade!47}87.7 & \cellcolor{kvshade!47}62.4 & \cellcolor{kvshade!65}\textit{86.5} & \cellcolor{kvshade!82}\textit{68.8} & \cellcolor{kvshade!30}\textit{73.3} & \cellcolor{kvshade!30}\textit{38.6} & \cellcolor{kvshade!47}78.8 & \cellcolor{kvshade!47}58.3 \\
\hw{HM} + EA & $\triangle$ &  & \cellcolor{kvshade!65}89.0 & \cellcolor{kvshade!100}\textbf{69.1} & \cellcolor{kvshade!65}88.0 & \cellcolor{kvshade!65}62.8 & \cellcolor{kvshade!47}84.9 & \cellcolor{kvshade!65}68.4 & \cellcolor{kvshade!65}92.0 & \cellcolor{kvshade!65}77.9 & \cellcolor{kvshade!82}80.5 & \cellcolor{kvshade!82}61.1 \\
\hw{HM} + Guide & $\triangle$ & $\bullet$ & \cellcolor{kvshade!82}89.1 & \cellcolor{kvshade!82}67.9 & \cellcolor{kvshade!82}89.9 & \cellcolor{kvshade!82}68.3 & \cellcolor{kvshade!82}87.7 & \cellcolor{kvshade!100}\textbf{72.9} & \cellcolor{kvshade!82}93.1 & \cellcolor{kvshade!82}78.7 & \cellcolor{kvshade!65}80.2 & \cellcolor{kvshade!65}60.1 \\
\midrule
Uncompressed &  &  & \multicolumn{2}{c}{93.9} & \multicolumn{2}{c}{94.6} & \multicolumn{2}{c}{92.1} & \multicolumn{2}{c}{95.6} & \multicolumn{2}{c}{89.8} \\
\bottomrule
\end{tabular}
\endgroup

\end{table}

\begin{table}[htbp]
\centering
\caption{\textbf{On LongBench, LU-KV + EA and HM + EA differ by at most 2.5 points over eight columns.} Unweighted mean over the 16 English LongBench subsets at their full official split (3750 documents), each scored by its official metric, one run per cell, for four targets at 8$\times$ and 16$\times$ compression; phi-4 was not run. No interval is given, since the mean mixes metrics. Methods, shading, cost symbols and italics as in \cref{tab:kv_ruler_4096}; the uncompressed row is not ranked.}

\label{tab:kv_longbench}
\begingroup
\footnotesize
\setlength{\tabcolsep}{1.6pt}
\ifdefined\kvshadecolw\else\newlength{\kvshadecolw}\fi
\settowidth{\kvshadecolw}{\footnotesize\bfseries Llama-3.1-8B}
\addtolength{\kvshadecolw}{-2\tabcolsep}
\divide\kvshadecolw by 2
\ifdefined\kvshadecolwfl\else\newlength{\kvshadecolwfl}\fi
\settowidth{\kvshadecolwfl}{\footnotesize\bfseries 42.1}
\ifdim\kvshadecolw<\kvshadecolwfl\setlength{\kvshadecolw}{\kvshadecolwfl}\fi
\begin{tabular}{>{\raggedright\arraybackslash}m{0.9in}cc>{\centering\arraybackslash}m{\kvshadecolw}>{\centering\arraybackslash}m{\kvshadecolw}>{\centering\arraybackslash}m{\kvshadecolw}>{\centering\arraybackslash}m{\kvshadecolw}>{\centering\arraybackslash}m{\kvshadecolw}>{\centering\arraybackslash}m{\kvshadecolw}>{\centering\arraybackslash}m{\kvshadecolw}>{\centering\arraybackslash}m{\kvshadecolw}}
\toprule
 & \multicolumn{2}{c}{\textbf{cost}} & \multicolumn{2}{c}{\textbf{Qwen3-8B}} & \multicolumn{2}{c}{\textbf{Llama-3.1-8B}} & \multicolumn{2}{c}{\textbf{Qwen3-32B}} & \multicolumn{2}{c}{\textbf{Mistral-7B}} \\
\cmidrule(lr){2-3}\cmidrule(lr){4-5}\cmidrule(lr){6-7}\cmidrule(lr){8-9}\cmidrule(lr){10-11}
 & \scriptsize offline & \scriptsize online & 8$\times$ & 16$\times$ & 8$\times$ & 16$\times$ & 8$\times$ & 16$\times$ & 8$\times$ & 16$\times$ \\
\midrule
AdaKV + EA &  &  & \cellcolor{kvshade!30}42.1 & \cellcolor{kvshade!47}33.1 & \cellcolor{kvshade!12}36.4 & \cellcolor{kvshade!12}28.4 & \cellcolor{kvshade!30}46.0 & \cellcolor{kvshade!30}39.2 & \cellcolor{kvshade!30}31.0 & \cellcolor{kvshade!30}27.6 \\
AdaKV + KVzap & $\triangle\triangle$ &  & \cellcolor{kvshade!12}34.0 & \cellcolor{kvshade!12}23.9 & \cellcolor{kvshade!30}42.0 & \cellcolor{kvshade!47}34.4 & \cellcolor{kvshade!12}36.8 & \cellcolor{kvshade!12}19.0 & \cellcolor{kvshade!12}\textit{28.5} & \cellcolor{kvshade!12}\textit{20.0} \\
KVzip &  & $\bullet\bullet$ & \cellcolor{kvshade!47}44.7 & \cellcolor{kvshade!30}28.0 & \cellcolor{kvshade!47}44.5 & \cellcolor{kvshade!30}33.8 & \cellcolor{kvshade!100}\textbf{50.7} & \cellcolor{kvshade!100}\textbf{45.2} & \cellcolor{kvshade!100}\textbf{43.8} & \cellcolor{kvshade!65}38.2 \\
LU-KV + EA & $\triangle$ &  & \cellcolor{kvshade!100}\textit{\textbf{46.3}} & \cellcolor{kvshade!100}\textit{\textbf{41.1}} & \cellcolor{kvshade!100}\textbf{45.8} & \cellcolor{kvshade!100}\textbf{43.0} & \cellcolor{kvshade!47}\textit{48.0} & \cellcolor{kvshade!47}\textit{43.4} & \cellcolor{kvshade!82}42.6 & \cellcolor{kvshade!82}38.8 \\
\hw{HM} + EA & $\triangle$ &  & \cellcolor{kvshade!65}46.2 & \cellcolor{kvshade!82}39.9 & \cellcolor{kvshade!82}45.6 & \cellcolor{kvshade!82}40.5 & \cellcolor{kvshade!82}49.9 & \cellcolor{kvshade!82}44.8 & \cellcolor{kvshade!47}42.3 & \cellcolor{kvshade!100}\textbf{39.5} \\
\hw{HM} + Guide & $\triangle$ & $\bullet$ & \cellcolor{kvshade!82}46.3 & \cellcolor{kvshade!65}38.0 & \cellcolor{kvshade!65}44.9 & \cellcolor{kvshade!65}37.1 & \cellcolor{kvshade!65}49.8 & \cellcolor{kvshade!65}44.3 & \cellcolor{kvshade!65}42.3 & \cellcolor{kvshade!47}36.4 \\
\midrule
Uncompressed &  &  & \multicolumn{2}{c}{47.3} & \multicolumn{2}{c}{45.9} & \multicolumn{2}{c}{50.9} & \multicolumn{2}{c}{42.9} \\
\bottomrule
\end{tabular}
\endgroup

\end{table}

\subsection{The guided score}
\label[appendix]{app:kv_details:guide}

The guided score keeps EA and raises the tokens that a small guide model, Qwen2.5-0.5B-Instruct (0.5B parameters), marks as important. The guide reads the context the target compresses, without the question, once on the reading queries and once in a reconstruction pass. For each guide head it takes two profiles over positions: the attention each token receives, summed over the reading queries without value weights, and the largest attention it receives from the reconstruction queries. Note that neither profile is strictly the base \cref{defn:marginalAttention}: the reading profile is the marginal attention with $\wt{h}{s}{t} \equiv 1$, without the value weight (the factor $1/n$ does not change ranks), and the reconstruction profile takes the maximum over queries instead of the mean. Each profile is carried to the target's tokens through the characters they cover and ranked within the document. A linear map, fitted once per target by ridge regression on 48 documents from the same Nemotron sample as the budget (\cref{app:kv_details:budget}), turns these profiles into a prediction $\hat g_{h,s}(X)$ of the KVzip+ log score for every key-value head $h$ of the target. The prediction tilts EA head by head:
\begin{equation}
    \label{eq:guide_tilt}
    \gamma_{h,s}(X) = \gamma^{\mathrm{EA}}_{h,s}(X)\,\bigl(1 + \lambda\, g_{h,s}(X)\bigr),
    \qquad
    g_{h,s}(X) = \frac{\hat g_{h,s}(X) - \min_{s'} \hat g_{h,s'}(X)}{\max_{s'} \hat g_{h,s'}(X) - \min_{s'} \hat g_{h,s'}(X)},
\end{equation}
where $\gamma^{\mathrm{EA}}_{h,s}(X)$ is the EA score of entry $(h,s)$, $g_{h,s}(X)$ rescales the prediction to $[0,1]$ within each head, and $\lambda = 2$ sets the strength of the tilt; $\lambda = 0$ gives EA. The budget $\hw{\beta_h}$ is the \headmass{} table and the tilt leaves it unchanged.

\FloatBarrier

\subsection{Per-task breakdown}
\label{app:kv_details:per_task}

Each figure in this subsection shows one benchmark at one compression ratio on one target: RULER at context 4096 or 8192, or LongBench. Each group of bars is one task, or one subset for LongBench, and the bars in a group are the methods of \cref{tab:kv_ruler_4096}, in the same order and under the same names: AdaKV~+~EA, AdaKV~+~KVzap, KVzip, LU-KV~+~EA, \hw{HM}~+~EA and \hw{HM}~+~Guide. The black line is the uncompressed model, so the distance from a bar down to the line is what a method loses on that task. The figures show which tasks make the differences between methods, which a single aggregate score hides. The aggregate scores in the main table (\cref{tab:kv_ruler_4096}) and in \cref{tab:kv_ruler_8192,tab:kv_longbench} are the means of these per-task scores.

\begin{figure}[h]
\centering
\includegraphics[width=\linewidth]{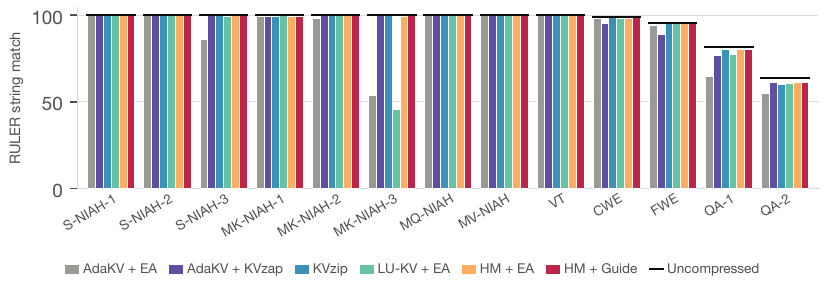}
\input{figures/main/kv_per_task_ruler4096_4x_qwen3-8b_caption.tex}
\label{fig:kv_per_task_ruler4096_4x_qwen3-8b}
\end{figure}

\begin{figure}[h]
\centering
\includegraphics[width=\linewidth]{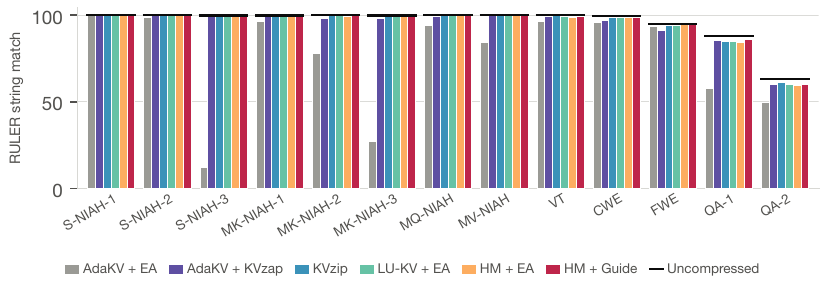}
\input{figures/main/kv_per_task_ruler4096_4x_llama-31-8b_caption.tex}
\label{fig:kv_per_task_ruler4096_4x_llama-31-8b}
\end{figure}

\begin{figure}[h]
\centering
\includegraphics[width=\linewidth]{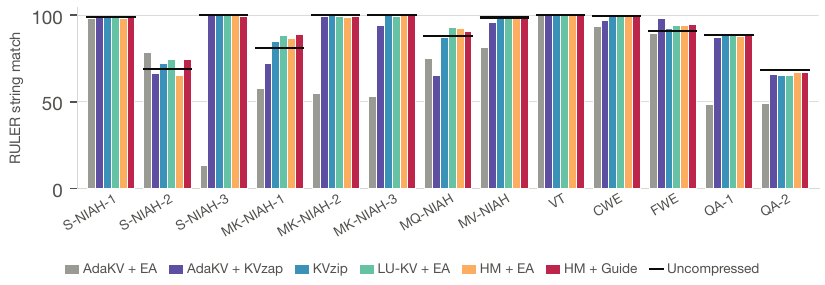}
\input{figures/main/kv_per_task_ruler4096_4x_phi-4_caption.tex}
\label{fig:kv_per_task_ruler4096_4x_phi-4}
\end{figure}

\begin{figure}[h]
\centering
\includegraphics[width=\linewidth]{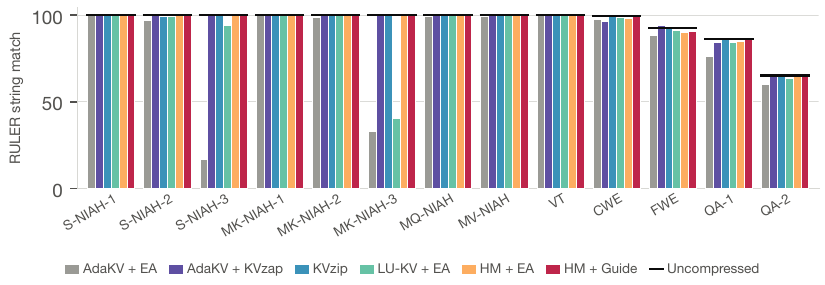}
\input{figures/main/kv_per_task_ruler4096_4x_qwen3-32b_caption.tex}
\label{fig:kv_per_task_ruler4096_4x_qwen3-32b}
\end{figure}

\begin{figure}[h]
\centering
\includegraphics[width=\linewidth]{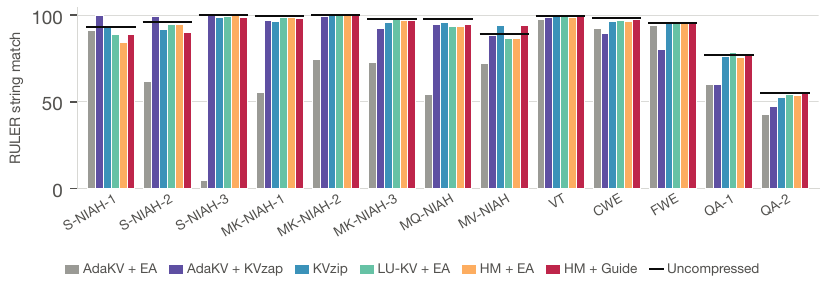}
\input{figures/main/kv_per_task_ruler4096_4x_mistral-7b_caption.tex}
\label{fig:kv_per_task_ruler4096_4x_mistral-7b}
\end{figure}

\begin{figure}[h]
\centering
\includegraphics[width=\linewidth]{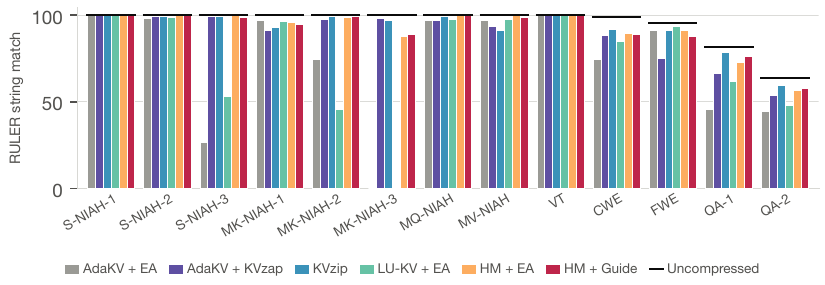}
\input{figures/main/kv_per_task_ruler4096_8x_qwen3-8b_caption.tex}
\label{fig:kv_per_task_ruler4096_8x_qwen3-8b}
\end{figure}

\begin{figure}[h]
\centering
\includegraphics[width=\linewidth]{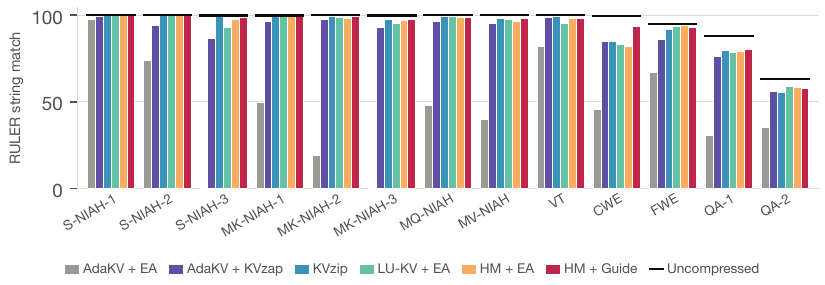}
\input{figures/main/kv_per_task_ruler4096_8x_llama-31-8b_caption.tex}
\label{fig:kv_per_task_ruler4096_8x_llama-31-8b}
\end{figure}

\begin{figure}[h]
\centering
\includegraphics[width=\linewidth]{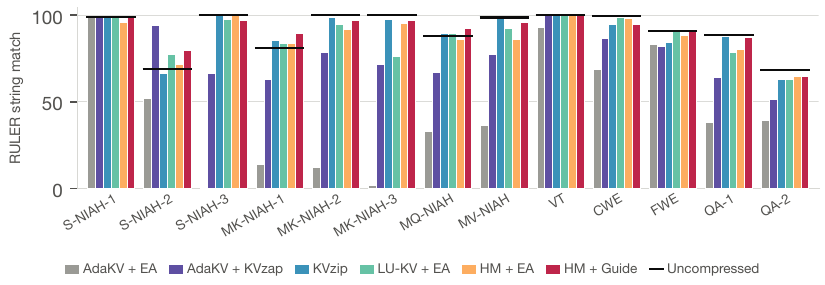}
\input{figures/main/kv_per_task_ruler4096_8x_phi-4_caption.tex}
\label{fig:kv_per_task_ruler4096_8x_phi-4}
\end{figure}

\begin{figure}[h]
\centering
\includegraphics[width=\linewidth]{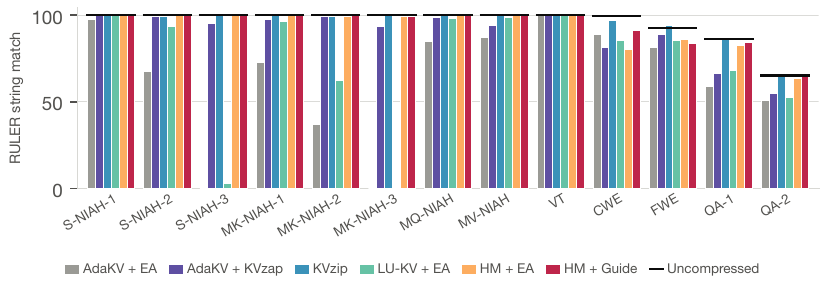}
\input{figures/main/kv_per_task_ruler4096_8x_qwen3-32b_caption.tex}
\label{fig:kv_per_task_ruler4096_8x_qwen3-32b}
\end{figure}

\begin{figure}[h]
\centering
\includegraphics[width=\linewidth]{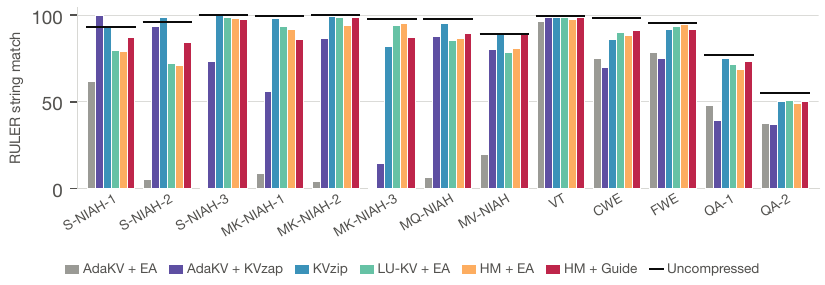}
\input{figures/main/kv_per_task_ruler4096_8x_mistral-7b_caption.tex}
\label{fig:kv_per_task_ruler4096_8x_mistral-7b}
\end{figure}

\begin{figure}[h]
\centering
\includegraphics[width=\linewidth]{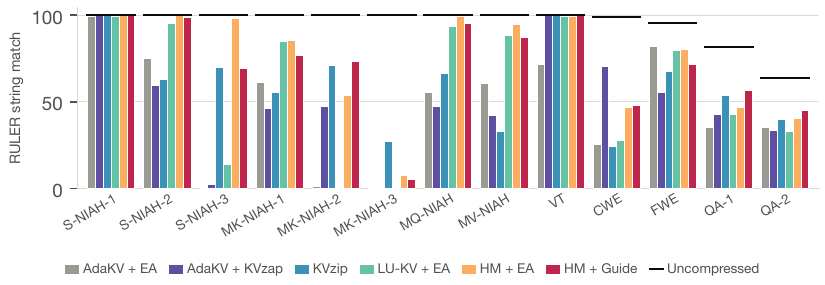}
\input{figures/main/kv_per_task_ruler4096_16x_qwen3-8b_caption.tex}
\label{fig:kv_per_task_ruler4096_16x_qwen3-8b}
\end{figure}

\begin{figure}[h]
\centering
\includegraphics[width=\linewidth]{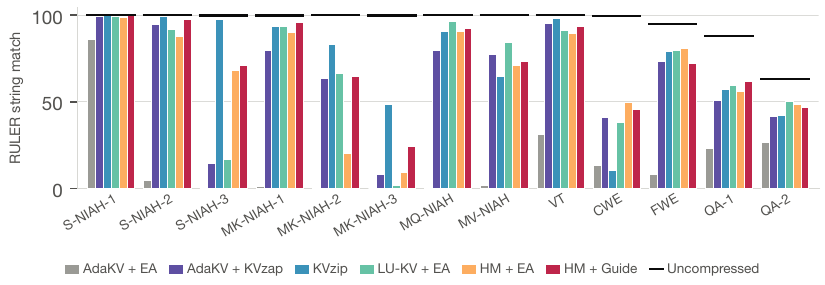}
\input{figures/main/kv_per_task_ruler4096_16x_llama-31-8b_caption.tex}
\label{fig:kv_per_task_ruler4096_16x_llama-31-8b}
\end{figure}

\begin{figure}[h]
\centering
\includegraphics[width=\linewidth]{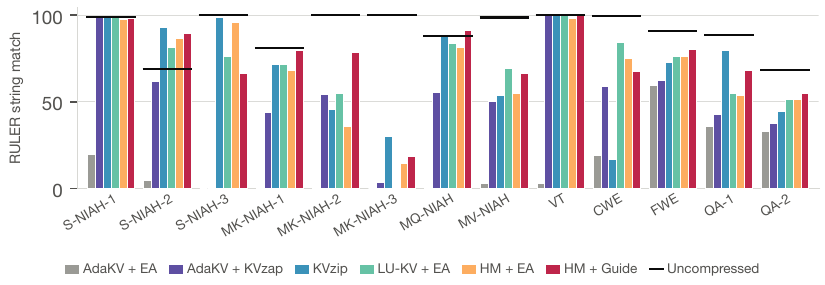}
\input{figures/main/kv_per_task_ruler4096_16x_phi-4_caption.tex}
\label{fig:kv_per_task_ruler4096_16x_phi-4}
\end{figure}

\begin{figure}[h]
\centering
\includegraphics[width=\linewidth]{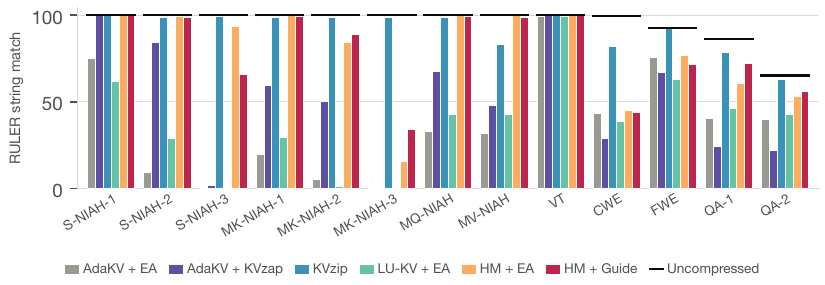}
\input{figures/main/kv_per_task_ruler4096_16x_qwen3-32b_caption.tex}
\label{fig:kv_per_task_ruler4096_16x_qwen3-32b}
\end{figure}

\begin{figure}[h]
\centering
\includegraphics[width=\linewidth]{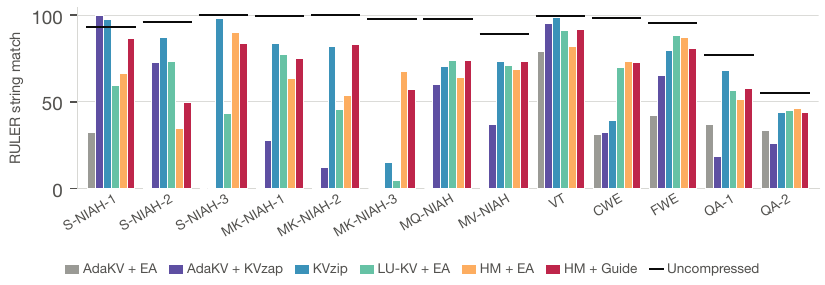}
\input{figures/main/kv_per_task_ruler4096_16x_mistral-7b_caption.tex}
\label{fig:kv_per_task_ruler4096_16x_mistral-7b}
\end{figure}

\begin{figure}[h]
\centering
\includegraphics[width=\linewidth]{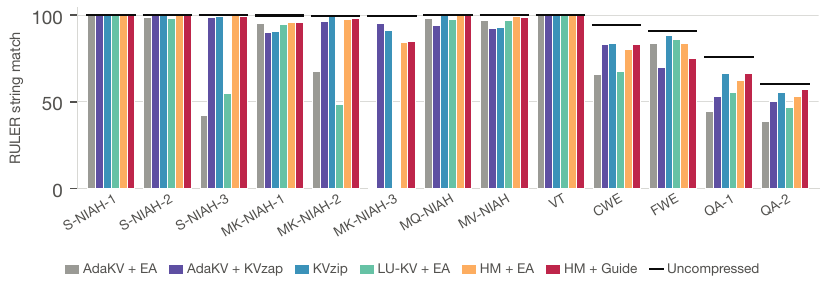}
\input{figures/main/kv_per_task_ruler8192_8x_qwen3-8b_caption.tex}
\label{fig:kv_per_task_ruler8192_8x_qwen3-8b}
\end{figure}

\begin{figure}[h]
\centering
\includegraphics[width=\linewidth]{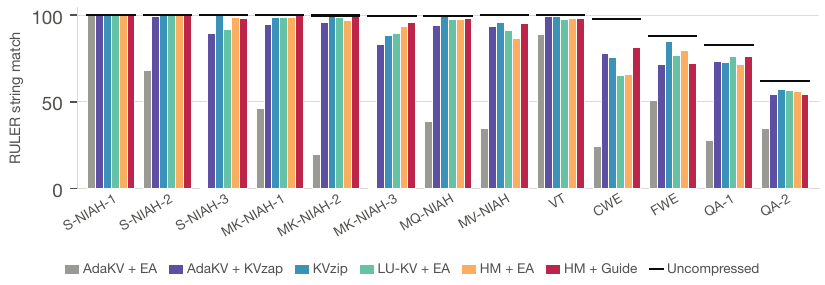}
\input{figures/main/kv_per_task_ruler8192_8x_llama-31-8b_caption.tex}
\label{fig:kv_per_task_ruler8192_8x_llama-31-8b}
\end{figure}

\begin{figure}[h]
\centering
\includegraphics[width=\linewidth]{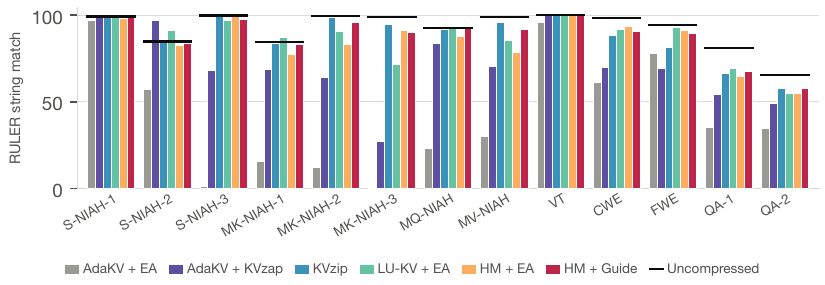}
\input{figures/main/kv_per_task_ruler8192_8x_phi-4_caption.tex}
\label{fig:kv_per_task_ruler8192_8x_phi-4}
\end{figure}

\begin{figure}[h]
\centering
\includegraphics[width=\linewidth]{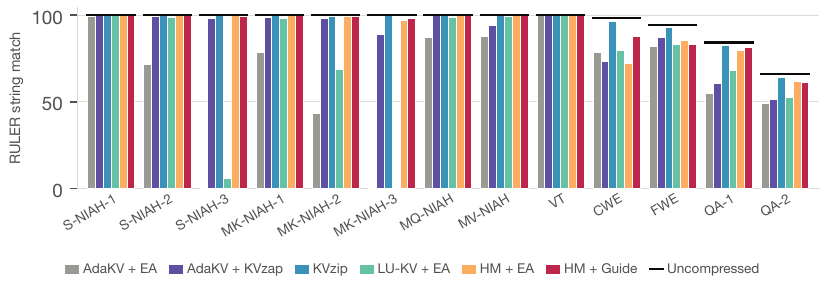}
\input{figures/main/kv_per_task_ruler8192_8x_qwen3-32b_caption.tex}
\label{fig:kv_per_task_ruler8192_8x_qwen3-32b}
\end{figure}

\begin{figure}[h]
\centering
\includegraphics[width=\linewidth]{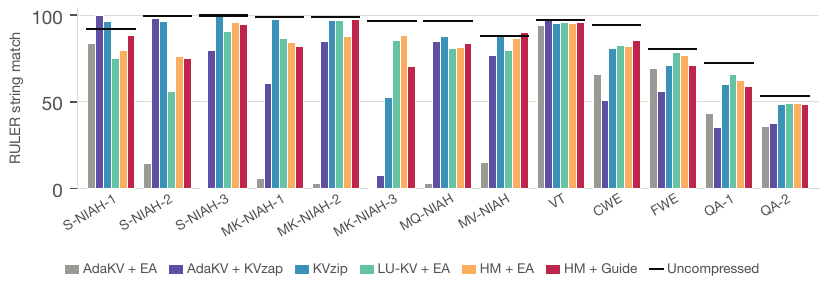}
\input{figures/main/kv_per_task_ruler8192_8x_mistral-7b_caption.tex}
\label{fig:kv_per_task_ruler8192_8x_mistral-7b}
\end{figure}

\begin{figure}[h]
\centering
\includegraphics[width=\linewidth]{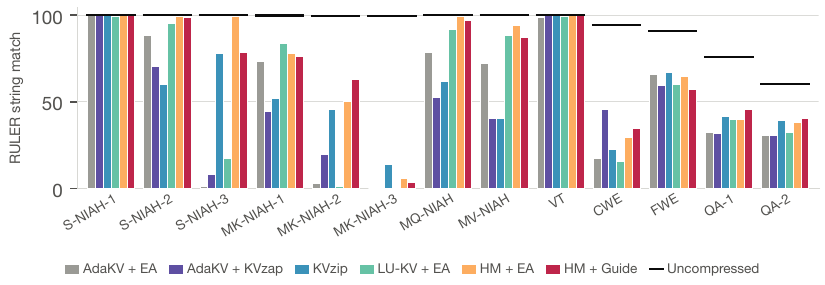}
\input{figures/main/kv_per_task_ruler8192_16x_qwen3-8b_caption.tex}
\label{fig:kv_per_task_ruler8192_16x_qwen3-8b}
\end{figure}

\begin{figure}[h]
\centering
\includegraphics[width=\linewidth]{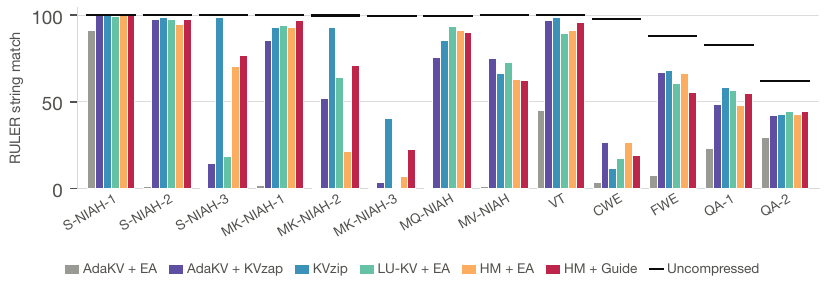}
\input{figures/main/kv_per_task_ruler8192_16x_llama-31-8b_caption.tex}
\label{fig:kv_per_task_ruler8192_16x_llama-31-8b}
\end{figure}

\begin{figure}[h]
\centering
\includegraphics[width=\linewidth]{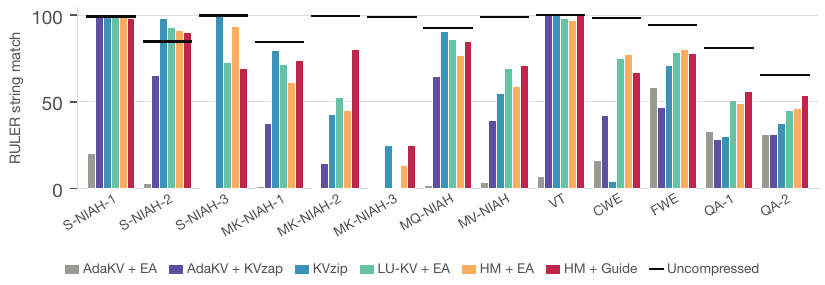}
\input{figures/main/kv_per_task_ruler8192_16x_phi-4_caption.tex}
\label{fig:kv_per_task_ruler8192_16x_phi-4}
\end{figure}

\begin{figure}[h]
\centering
\includegraphics[width=\linewidth]{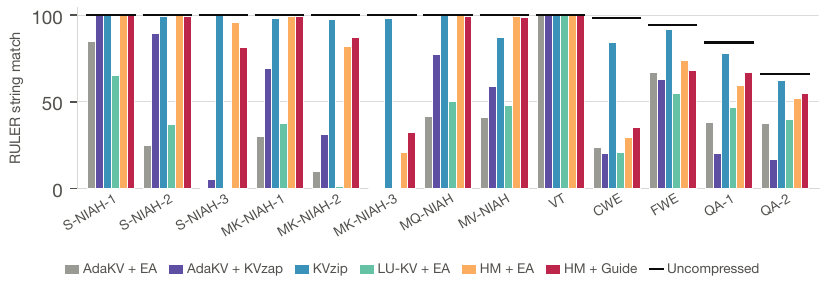}
\input{figures/main/kv_per_task_ruler8192_16x_qwen3-32b_caption.tex}
\label{fig:kv_per_task_ruler8192_16x_qwen3-32b}
\end{figure}

\begin{figure}[h]
\centering
\includegraphics[width=\linewidth]{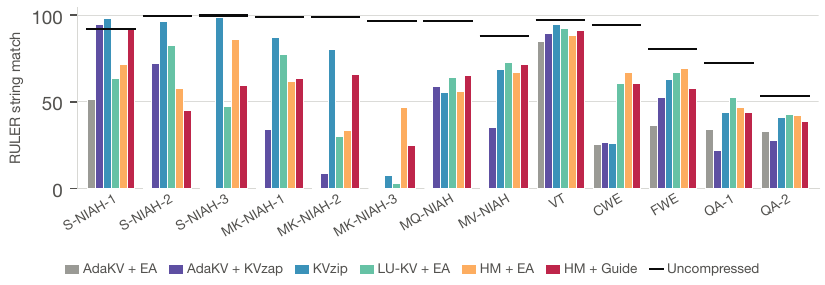}
\input{figures/main/kv_per_task_ruler8192_16x_mistral-7b_caption.tex}
\label{fig:kv_per_task_ruler8192_16x_mistral-7b}
\end{figure}

\begin{figure}[h]
\centering
\includegraphics[width=\linewidth]{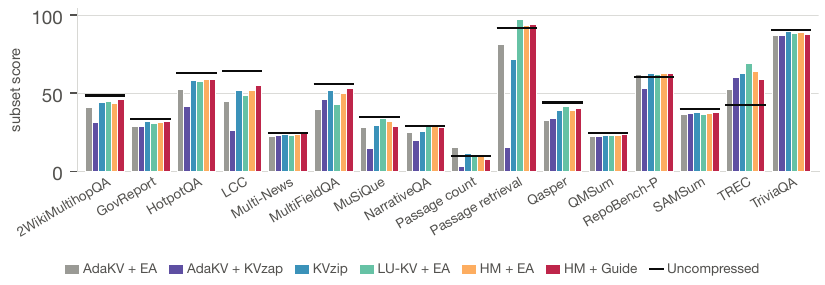}
\input{figures/main/kv_per_task_longbench_8x_qwen3-8b_caption.tex}
\label{fig:kv_per_task_longbench_8x_qwen3-8b}
\end{figure}

\begin{figure}[h]
\centering
\includegraphics[width=\linewidth]{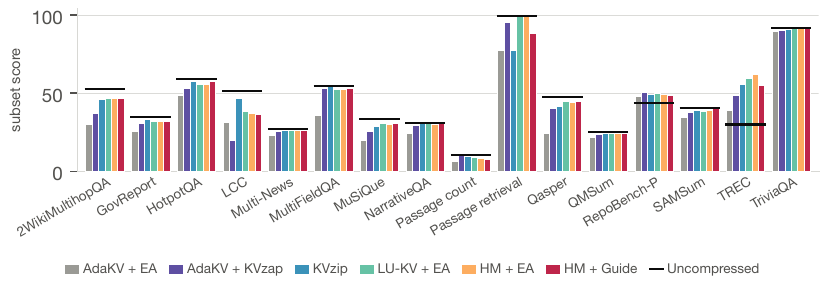}
\input{figures/main/kv_per_task_longbench_8x_llama-31-8b_caption.tex}
\label{fig:kv_per_task_longbench_8x_llama-31-8b}
\end{figure}

\begin{figure}[h]
\centering
\includegraphics[width=\linewidth]{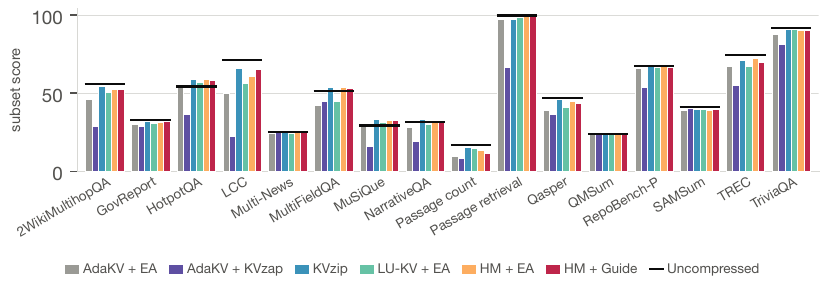}
\input{figures/main/kv_per_task_longbench_8x_qwen3-32b_caption.tex}
\label{fig:kv_per_task_longbench_8x_qwen3-32b}
\end{figure}

\begin{figure}[h]
\centering
\includegraphics[width=\linewidth]{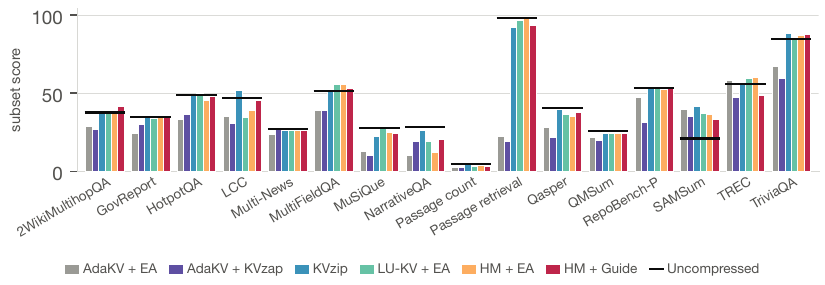}
\input{figures/main/kv_per_task_longbench_8x_mistral-7b_caption.tex}
\label{fig:kv_per_task_longbench_8x_mistral-7b}
\end{figure}

\begin{figure}[h]
\centering
\includegraphics[width=\linewidth]{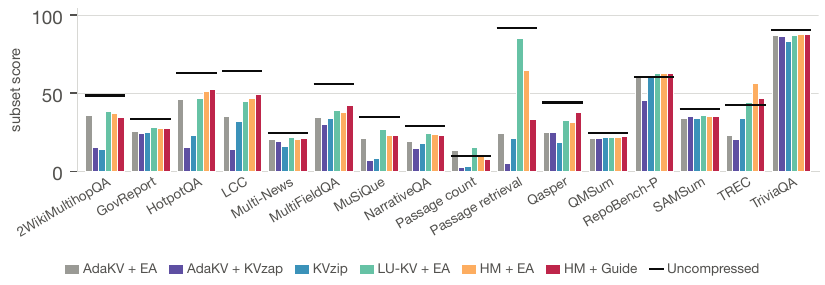}
\input{figures/main/kv_per_task_longbench_16x_qwen3-8b_caption.tex}
\label{fig:kv_per_task_longbench_16x_qwen3-8b}
\end{figure}

\begin{figure}[h]
\centering
\includegraphics[width=\linewidth]{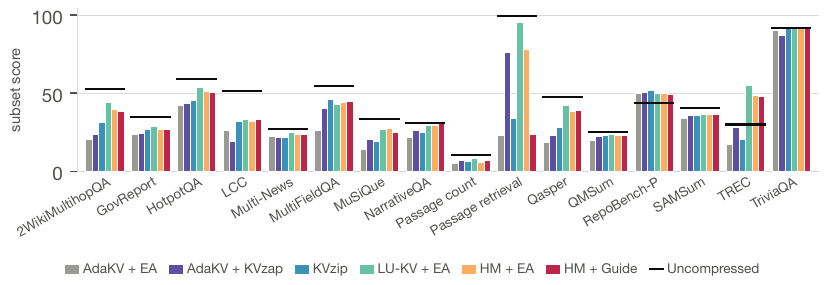}
\input{figures/main/kv_per_task_longbench_16x_llama-31-8b_caption.tex}
\label{fig:kv_per_task_longbench_16x_llama-31-8b}
\end{figure}

\begin{figure}[h]
\centering
\includegraphics[width=\linewidth]{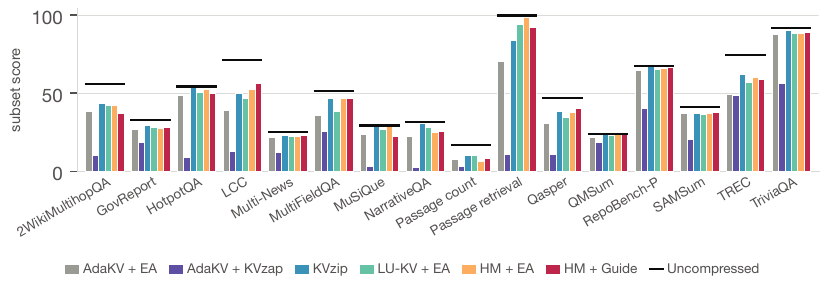}
\input{figures/main/kv_per_task_longbench_16x_qwen3-32b_caption.tex}
\label{fig:kv_per_task_longbench_16x_qwen3-32b}
\end{figure}

\begin{figure}[h]
\centering
\includegraphics[width=\linewidth]{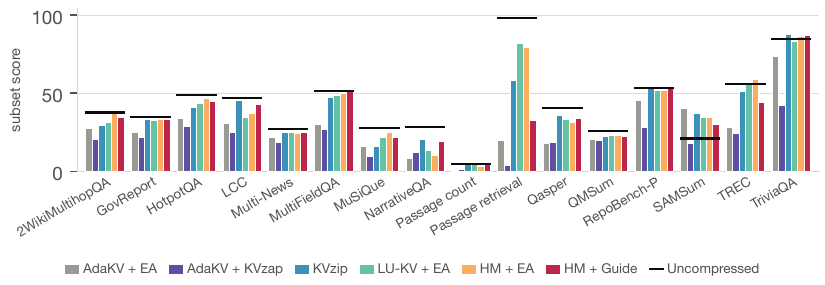}
\input{figures/main/kv_per_task_longbench_16x_mistral-7b_caption.tex}
\label{fig:kv_per_task_longbench_16x_mistral-7b}
\end{figure}

\FloatBarrier

\section{Proofs}
\label{app:proofs}

Throughout this appendix we retain the notation of the main text. Recall that $n$ refers to the length of $X$ (the number of input positions, indexed by $s$ and output positions, indexed by $t$), that $J^{(s)}$ without argument refers to $J^{(s)}(X)$ and $J$ refers to $J(X)$, and that $\eta := \inf_{0 < r < R} \left( \frac{2\delta}{r} + Mr \right)$.

\textbf{Remark.} In practice, the next-token distribution is blind to a fixed constant being added to each logit, but this is trivially ignored by applying the theorem to the centered logits.

\subsection{Assumptions}
\label{app:proofs:assumptions}

\Cref{thm:jacobianControl} relies on three assumptions, which we state formally here. The first, smoothness, is all that \cref{lem:jacobianBlocks} needs.

\begin{assumption}[Jacobian Smoothness]
    \label{assump:jacobianSmoothness}
    Each of $f$ and $f'$ is continuously differentiable in each input position, and there are constants $M \geq 0$ and $R > 0$ such that
    \begin{equation*}
        \left\lVert \left[ J(X^{(s) + r\mathbf{u}}) \right]_t - \left[ J(X) \right]_t \right\rVert_F, \left\lVert \left[ J'(X^{(s) + r\mathbf{u}}) \right]_t - \left[ J'(X) \right]_t \right\rVert_F \leq Mr
    \end{equation*}
    $\mathbf{P}$-almost everywhere in $X$, for all $0 \leq r < R$ and unit directions $\mathbf{u}$, and for all $s \leq t$ over all $t$.
\end{assumption}

\Cref{assump:jacobianSmoothness} is a fairly standard assumption on the curvature of the model $f$. Jacobian-vector products provide local diagnostics of this behaviour. Empirically, larger models show lower relative local sensitivity on the sampled inputs (\cref{app:theory_measurements}); these measurements do not establish a uniform bound on $M$.

The other two constrain the dependence structure of the Jacobian. For $k \in \mathbb{N}$, write $\Psi_k(X) = \frac{1}{n\sqrt d} \sum_{(s,t):\, s \le t-k} \lVert [J^{(s)}]_t \rVert_F$ for the lag-$k$ tail of the Jacobian block norms of $f$, and $\Psi'_k(X)$ for the same quantity for $f'$.

\begin{assumption}[Preceding-Token Dependence]
    \label{assump:preTokenDependence}
    There is a constant $c > 0$ such that
    \begin{equation*}
        \left\lVert \left[ J^{(t)}(X) \right]_t \right\rVert_F^2, \left\lVert \left[ J'^{(t)}(X) \right]_t \right\rVert_F^2 \geq cd
    \end{equation*}
    $\mathbf{P}$-almost everywhere in $X$, for all output positions $t$.
\end{assumption}

\Cref{assump:preTokenDependence} states that every prediction depends, by a margin bounded away from zero, on the last token it reads, the one at its own position. We find this block above zero at every sampled position of every model we examine (\cref{app:theory_measurements:constants}).

\begin{assumption}[Lag Decay]
    \label{assump:lagDecay}
    There is a non-increasing function $\psi : \mathbb{N} \to [0, \infty)$, independent of $n$ and $d$, with $\psi(k) \to 0$ as $k \to \infty$, such that
    \begin{equation*}
        \Psi_k(X), \Psi'_k(X) \leq \psi(k)
    \end{equation*}
    $\mathbf{P}$-almost everywhere in $X$, for all $k \in \mathbb{N}$.
\end{assumption}

\Cref{assump:lagDecay} is the one that depends most on the data: it states that the dependence of a prediction on earlier tokens falls toward zero with their distance, at a rate that does not depend on the length of the sequence. We observe a clear but slow decay, close to a power law, and larger models decay more slowly (\cref{app:theory_measurements:constants}). The assumption is needed only because of our choice of units: an alternative proof with squared block norms gives a bound independent of $n$ without it. Whether a milder assumption suffices for this statistic, and whether its normalisation is the right one, is an open question.

\subsection{Proofs}

\begin{lemma}[Near-Identifiability of the Jacobian Blocks]
    \label{lem:jacobianBlocks}
    Let $\mathbf{P}(X)$ be a distribution supported on a space of canonical input sequences $\mathcal{X}$, where each element of the sequence is an embedding in $\mathbb{R}^d$ and sequences may be of any length $n$. Let $f, f'$ be the functions of two large language models mapping sequences of input embeddings to sequences of output logit vectors, each of whose Jacobians is smooth in each input position in the sense of \cref{assump:jacobianSmoothness}, with constants $M$ and $R$ as given there. Suppose that the next-token logits implied by $f$ and $f'$ are similar for each output position, in the sense that
    \begin{equation*}
        \left\lVert f_t(X^{(s) + r\mathbf{u}}) - f_t'(X^{(s) + r\mathbf{u}}) \right\rVert \leq \delta
    \end{equation*}
    $\mathbf{P}$-almost everywhere in $X$, for some $\delta \geq 0$, for all $0 \leq r < R$ and unit directions $\mathbf{u}$, and for all $s \leq t$ over all $t$. Then, writing $\eta := \inf_{0 < r < R} \left( \frac{2\delta}{r} + Mr \right)$, we have
    \begin{equation*}
        \left\lVert \left[ J^{(s)}(X) \right]_t - \left[ J'^{(s)}(X) \right]_t \right\rVert_F \leq \sqrt{d} \, \eta
    \end{equation*}
    $\mathbf{P}$-almost everywhere in $X$, for all $s \leq t$ over all $t$. Moreover, $\eta \leq \frac{4\delta}{R} + \frac{MR}{2}$, and $\eta = 2\sqrt{2M\delta}$ whenever $2\delta \leq MR^2$; in particular, $\eta$ depends on neither $n$ nor $d$, and $\eta \to 0$ as $M, \delta \to 0$.
\end{lemma}

\begin{proof}[Proof of \cref{lem:jacobianBlocks}]
    \textbf{Claim.} For all $0 \leq r < R$, unit vectors $\mathbf{u}$, and positions $s \leq t$, we have $\lVert f_t(X^{(s) + r\mathbf{u}}) - f_t(X) - r \left[ J^{(s)} \right]_t \mathbf{u} \rVert \leq \frac{1}{2} Mr^2$. The same bound holds for $f'$ and $[J'^{(s)}]_t$.

    \textbf{Proof of claim.} Write $g(\rho) = f_t(X^{(s) + \rho \mathbf{u}})$ for $\rho \in [0, r]$. By \cref{assump:jacobianSmoothness}, $g$ is continuously differentiable, and by the chain rule $g'(\rho) = \left[ J^{(s)}(X^{(s) + \rho \mathbf{u}}) \right]_t \mathbf{u}$. By the fundamental theorem of calculus,
    \begin{align*}
        \left\lVert f_t(X^{(s) + r\mathbf{u}}) - f_t(X) - r \left[ J^{(s)} \right]_t \mathbf{u} \right\rVert &= \left\lVert g(r) - g(0) - r \left[ J^{(s)} \right]_t \mathbf{u} \right\rVert \\
        &= \left\lVert \int_0^r \left( \left[ J^{(s)}(X^{(s) + \rho \mathbf{u}}) \right]_t - \left[ J^{(s)} \right]_t \right) \mathbf{u} \, d\rho \right\rVert \\
        &\leq \int_0^r M\rho \, d\rho = \frac{1}{2} Mr^2
    \end{align*}
    where the last line follows from the triangle inequality for integrals together with \cref{assump:jacobianSmoothness}, using $\lVert A \mathbf{u} \rVert \leq \lVert A \rVert_{\mathrm{op}} \leq \lVert A \rVert_F$ for unit $\mathbf{u}$. The identical argument applies to $f'$ and $J'$.

    \textbf{Claim.} For all positions $s \leq t$ and every unit vector $\mathbf{u}$, we have $\left\lVert \left( \left[ J^{(s)} \right]_t - \left[ J'^{(s)} \right]_t \right) \mathbf{u} \right\rVert \leq \eta$.

    \textbf{Proof of claim.} Let $0 < r < R$ be arbitrary. By the previous claim, we have
    \begin{equation*}
        f_t(X^{(s) + r\mathbf{u}}) - f_t(X) - r \left[ J^{(s)} \right]_t \mathbf{u} - \zeta = 0
    \end{equation*}
    for some $\lVert \zeta \rVert \leq \frac{1}{2} Mr^2$, and similarly for $f'$, $[J'^{(s)}]_t$, and $\zeta'$. Subtracting the two equations and re-arranging yields
    \begin{align*}
        \left\lVert r \left[ J^{(s)} \right]_t \mathbf{u} - r \left[ J'^{(s)} \right]_t \mathbf{u} \right\rVert &= \left\lVert f_t(X^{(s) + r \mathbf{u}}) - f'_t(X^{(s) + r\mathbf{u}}) - f_t(X) + f'_t(X) - \zeta + \zeta' \right\rVert \\
        &\leq 2\delta + Mr^2
    \end{align*}
    where the final bound follows by the triangle inequality and the near-agreement of the next-token logits hypothesized in the lemma, the latter applied at $r$ and at $0$. Dividing by $r$ and taking the infimum over $0 < r < R$ yields the claim.

    It remains to pass to the Frobenius norm and to estimate $\eta$. Write $A := \left[ J^{(s)} \right]_t - \left[ J'^{(s)} \right]_t \in \mathbb{R}^{V \times d}$. Taking the supremum over unit $\mathbf{u}$ in the previous claim gives $\lVert A \rVert_{\mathrm{op}} \leq \eta$, and since $\operatorname{rank}(A) \leq d$ we have $\lVert A \rVert_F^2 = \sum_i \sigma_i(A)^2 \leq d \lVert A \rVert_{\mathrm{op}}^2 \leq d\eta^2$. For the explicit estimates, taking $r = R/2$ in the definition of $\eta$ gives $\eta \leq \frac{4\delta}{R} + \frac{MR}{2}$, while the unconstrained minimizer of $r \mapsto \frac{2\delta}{r} + Mr$ over $r > 0$ is $r^\ast = \sqrt{2\delta/M}$, attaining the value $2\sqrt{2M\delta}$, which is admissible precisely when $r^\ast \leq R$. Both estimates depend only on $M, \delta$ and $R$, and hence on neither $n$ nor $d$.
\end{proof}

\begin{restatedtheorem}{thm:jacobianControl}[Near-Identifiability of the Per-Token Marginal Influence, formal]
    Let $\mathbf{P}(X)$ be a distribution supported on a space of canonical input sequences $\mathcal{X}$, where each element of the sequence is an embedding in $\mathbb{R}^d$ and sequences may be of any length $n$. Let $f, f'$ be the functions of two large language models mapping sequences of input embeddings to sequences of output logit vectors. Assume that (1) the prediction of the next token depends sufficiently on the immediately preceding token (\cref{assump:preTokenDependence}), (2) each model's Jacobian is smooth in each input position (\cref{assump:jacobianSmoothness}), and (3) the blockwise Jacobian norms of each model decay with lag (\cref{assump:lagDecay}), with constants $c$, $M$ and $R$ and envelope $\psi$ as given there. Suppose moreover that the next-token logits implied by $f$ and $f'$ are similar for each output position, in the sense that
    \begin{equation*}
        \left\lVert f_t(X^{(s) + r\mathbf{u}}) - f_t'(X^{(s) + r\mathbf{u}}) \right\rVert \leq \delta
    \end{equation*}
    $\mathbf{P}$-almost everywhere in $X$, for some $\delta \geq 0$, for all $0 \leq r < R$ and unit directions $\mathbf{u}$, and for all $s \leq t$ over all $t$. Then, we have
    \begin{equation*}
        \lVert \infl(X) - \infl'(X) \rVert_1 \leq \epsilon
    \end{equation*}
    $\mathbf{P}$-almost everywhere in $X$, where $\infl, \infl'$ are the \iw{per-token marginal influences} of $f$ and $f'$ respectively, and where $\epsilon$ depends on neither $n$ nor $d$, and $\epsilon \xrightarrow{M, \delta \to 0} 0$. Explicitly, $\epsilon := \min\{2, \frac{2}{\sqrt{c}} \inf_{k \in \mathbb{N}} (k\eta + 2\psi(k))\}$, which depends on $c$ and $\psi$.
\end{restatedtheorem}

\begin{proof}[Proof of \cref{thm:jacobianControl}]
    The hypotheses of the theorem include those of \cref{lem:jacobianBlocks}, which therefore gives $\lVert [ J^{(s)} ]_t - [ J'^{(s)} ]_t \rVert_F \leq \sqrt{d} \, \eta$ for all $s \leq t$.

    \textbf{Claim.} We have $Z(X) \geq n\sqrt{cd}$ and $Z'(X) \geq n\sqrt{cd}$; equivalently, $\Psi_0(X), \Psi'_0(X) \geq \sqrt{c}$. In particular, $\infl(X)$ and $\infl'(X)$ are well defined.

    \textbf{Proof of claim.} Every term of $Z(X) = \sum_{s \leq t} \lVert [J^{(s)}]_t \rVert_F$ is non-negative, so discarding all off-diagonal terms and applying \cref{assump:preTokenDependence} to each of the $n$ diagonal blocks gives
    \begin{equation*}
        Z(X) \geq \sum_{t=1}^n \left\lVert \left[ J^{(t)} \right]_t \right\rVert_F \geq n\sqrt{cd},
    \end{equation*}
    and identically for $f'$. Since $c > 0$, both normalizations are strictly positive.

    \textbf{Claim.} Let $F, F' \in [0, \infty)^n$ with $Z := \lVert F \rVert_1 > 0$ and $Z' := \lVert F' \rVert_1 > 0$. Then
    \begin{equation*}
        \left\lVert \frac{F}{Z} - \frac{F'}{Z'} \right\rVert_1 \leq \frac{2 \lVert F - F' \rVert_1}{\max(Z, Z')}.
    \end{equation*}

    \textbf{Proof of claim.} Decompose
    \begin{equation*}
        \frac{F}{Z} - \frac{F'}{Z'} = \frac{F - F'}{Z} + F' \left( \frac{1}{Z} - \frac{1}{Z'} \right).
    \end{equation*}
    The second term has $\ell^1$ norm $\lVert F' \rVert_1 \frac{\lvert Z' - Z \rvert}{ZZ'} = \frac{\lvert Z - Z' \rvert}{Z}$, and $\lvert Z - Z' \rvert = \bigl\lvert \lVert F \rVert_1 - \lVert F' \rVert_1 \bigr\rvert \leq \lVert F - F' \rVert_1$ by the reverse triangle inequality. Hence the total is at most $2 \lVert F - F' \rVert_1 / Z$. Exchanging the roles of $F$ and $F'$ gives the same bound with $Z'$ in place of $Z$, so we may take whichever of the two is larger.

    \textbf{Claim.} For every $k \in \mathbb{N}$, we have $\lVert F(X) - F'(X) \rVert_1 \leq n\sqrt{d} \left( k\eta + 2\psi(k) \right)$.

    \textbf{Proof of claim.} By the triangle inequality applied to each $F_s(X) - F'_s(X)$,
    \begin{equation*}
        \lVert F(X) - F'(X) \rVert_1 \leq \sum_{s \leq t} \Bigl\lvert \left\lVert \left[ J^{(s)} \right]_t \right\rVert_F - \left\lVert \left[ J'^{(s)} \right]_t \right\rVert_F \Bigr\rvert.
    \end{equation*}
    Partition $\{ (s,t) : s \leq t \}$ into the recent pairs $\mathcal{R}_k := \{ (s,t) : t - k < s \leq t \}$ and the distant pairs $\mathcal{D}_k := \{ (s,t) : s \leq t-k \}$. On $\mathcal{R}_k$, the reverse triangle inequality and \cref{lem:jacobianBlocks} give
    \begin{equation*}
        \sum_{(s,t) \in \mathcal{R}_k} \Bigl\lvert \left\lVert \left[ J^{(s)} \right]_t \right\rVert_F - \left\lVert \left[ J'^{(s)} \right]_t \right\rVert_F \Bigr\rvert \leq \sum_{(s,t) \in \mathcal{R}_k} \left\lVert \left[ J^{(s)} \right]_t - \left[ J'^{(s)} \right]_t \right\rVert_F \leq nk\sqrt{d} \, \eta,
    \end{equation*}
    since each output position $t$ contributes at most $k$ recent pairs, so $\lvert \mathcal{R}_k \rvert \leq nk$. On $\mathcal{D}_k$, we discard the cancellation via $\lvert a - b \rvert \leq a + b$ and recognize the two resulting sums as the lag-$k$ tails $\Psi_k(X)$ and $\Psi'_k(X)$, so that by \cref{assump:lagDecay},
    \begin{equation*}
        \sum_{(s,t) \in \mathcal{D}_k} \left( \left\lVert \left[ J^{(s)} \right]_t \right\rVert_F + \left\lVert \left[ J'^{(s)} \right]_t \right\rVert_F \right) = n\sqrt{d} \left( \Psi_k(X) + \Psi'_k(X) \right) \leq 2n\sqrt{d} \, \psi(k).
    \end{equation*}
    Adding the two contributions yields the claim. For $k = 0$ the set $\mathcal{R}_0$ is empty and the bound holds trivially.

    \textbf{Claim.} Define
    \begin{equation*}
        \epsilon := \min \left\{ 2, \ \frac{2}{\sqrt{c}} \inf_{k \in \mathbb{N}} \left( k\eta + 2\psi(k) \right) \right\}.
    \end{equation*}
    Then $\epsilon$ depends on neither $n$ nor $d$, and $\epsilon \to 0$ as $M, \delta \to 0$.

    \textbf{Proof of claim.} Independence of $n$ and $d$ is immediate, since $\eta$ depends only on $M, \delta, R$, the function $\psi$ is independent of $n$ and $d$ by \cref{assump:lagDecay}, and $c$ is a fixed constant. For the limit, fix $\varepsilon > 0$. As $\psi(k) \to 0$, choose $k_0$ with $4\psi(k_0)/\sqrt{c} < \varepsilon/2$. Holding $k_0$ fixed, we have $\eta \leq \frac{4\delta}{R} + \frac{MR}{2} \to 0$ as $M, \delta \to 0$, so $2k_0\eta/\sqrt{c} < \varepsilon/2$ for all sufficiently small $M, \delta$, whence $\epsilon \leq \frac{2}{\sqrt{c}}(k_0 \eta + 2\psi(k_0)) < \varepsilon$.

    Combining the three preceding claims, for every $k \in \mathbb{N}$ we have
    \begin{equation*}
        \lVert \infl(X) - \infl'(X) \rVert_1 \leq \frac{2 \lVert F(X) - F'(X) \rVert_1}{\max(Z(X), Z'(X))} \leq \frac{2 n \sqrt{d} \left( k\eta + 2\psi(k) \right)}{n \sqrt{cd}} = \frac{2}{\sqrt{c}} \left( k\eta + 2\psi(k) \right),
    \end{equation*}
    while $\lVert \infl(X) - \infl'(X) \rVert_1 \leq 2$ unconditionally, both $\infl(X)$ and $\infl'(X)$ being non-negative with unit $\ell^1$ norm. Taking the infimum over $k \in \mathbb{N}$ gives $\lVert \infl(X) - \infl'(X) \rVert_1 \leq \epsilon$.
\end{proof}

\end{document}